\documentclass[10pt,twocolumn,letterpaper]{article}

\usepackage[pagenumbers]{cvpr} 

\usepackage[dvipsnames]{xcolor}

\usepackage{makecell}
\usepackage{stfloats} 
\usepackage{soul} 
\usepackage{lipsum}
\usepackage{xcolor}
\usepackage{colortbl}

\usepackage{float}

\definecolor{gray}{rgb}{0.7,0.7,0.7}
\definecolor{lightblue}{rgb}{0.3, 0.5, 1}
\definecolor{lightgreen}{rgb}{0.1, 0.5, 0.1}
\definecolor{salmon}{rgb}{1, 0.3, 0.3}

\usepackage{graphicx}
\usepackage{amsmath}
\usepackage{amssymb}  
\usepackage{lipsum}
\usepackage{multirow}

\usepackage{amsthm}

\usepackage[ruled]{algorithm}
\usepackage{algpseudocode}

\usepackage{./rpm_packages/rpm_SIunits}
\usepackage{./rpm_packages/rpm_acronyms}
\usepackage{./rpm_packages/rpm_math}
\usepackage{./rpm_packages/rpm_misc}

\usepackage{subcaption}
\usepackage{multirow}
\usepackage{array,booktabs}
\usepackage{diagbox}
\usepackage{balance}
\usepackage[ruled]{algorithm}
\usepackage{algpseudocode}
\usepackage{slashbox}
\usepackage{array}
\newcolumntype{P}[1]{>{\centering\arraybackslash}p{#1}}
\usepackage{enumitem}
\usepackage{slashbox}

\definecolor{cvprblue}{rgb}{0.21,0.49,0.74}
\usepackage[pagebackref,breaklinks,colorlinks,allcolors=cvprblue]{hyperref}

\def\paperID{32039} 
\def\confName{CVPR}
\def\confYear{2026}

\title{TherA: Thermal-Aware Visual-Language Prompting\\ for Controllable RGB-to-Thermal Infrared Translation}

\author{Dong-Guw Lee$^*$\\
Seoul National University\\
{\tt\small donkeymouse@snu.ac.kr}
\and
Tai Hyoung Rhee$^*$\\
Seoul National University\\
{\tt\small williamrhee@snu.ac.kr}
\and
Hyunsoo Jang\\
Seoul National University\\
{\tt\small bronto3082@snu.ac.kr}
\and
Young-Sik Shin\\
Kyungpook National University\\
{\tt\small yshin86@knu.ac.kr}
\and
Ukcheol Shin\\
KENTECH\\
{\tt\small ushin@kentech.ac.kr}
\and
Ayoung Kim\\
Seoul National University\\
{\tt\small ayoungk@snu.ac.kr}
}

\begin{document}
\maketitle
\def\thefootnote{*}\footnotetext{Equal contribution.}

\begin{abstract}
Despite the inherent advantages of \ac{TIR} imaging, large-scale data collection and annotation remain a major bottleneck for \ac{TIR}-based perception.
A practical alternative is to synthesize pseudo-\ac{TIR} data via image translation; however, most RGB-to-\ac{TIR} approaches heavily rely on RGB-centric priors that overlook thermal physics, yielding implausible heat distributions. In this paper, we introduce TherA, a controllable RGB-to-\ac{TIR} translation framework that produces diverse and thermally plausible images at both scene and object level.
TherA couples TherA-VLM with a latent-diffusion-based translator. Given a single RGB image and a user-prompted condition pair, TherA-VLM yields a thermal-aware embedding that encodes scene, object, material, and heat-emission context reflecting the input scene-condition pair. Conditioning the diffusion model on this embedding enables realistic \ac{TIR} synthesis and fine-grained control across time of day, weather, and object state. Compared to other baselines, TherA achieves state-of-the-art translation performance, demonstrating improved zero-shot translation performance up to 33\% increase averaged across all metrics.

\end{abstract}
   
\section{Introduction}

\begin{figure}[t]
  \centering
  \begin{subfigure}{\columnwidth}
    \centering
    \includegraphics[width=\columnwidth]{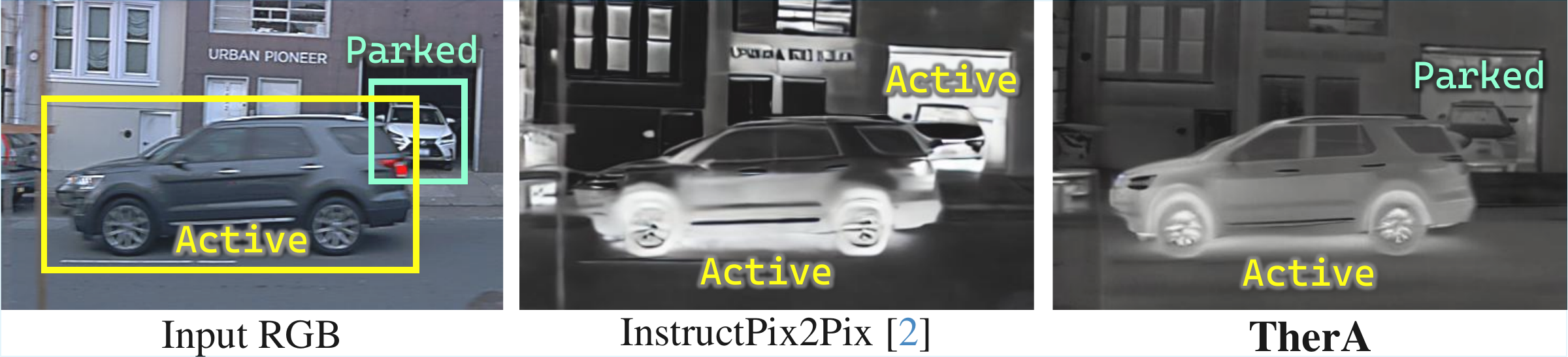}
    \caption{Thermal-aware RGB-to-TIR translation. }
    \label{thermal-aware-fig1}
  \end{subfigure}
  \vspace{0.3em}
  \begin{subfigure}{\columnwidth}
    \centering
    \includegraphics[width=\columnwidth]{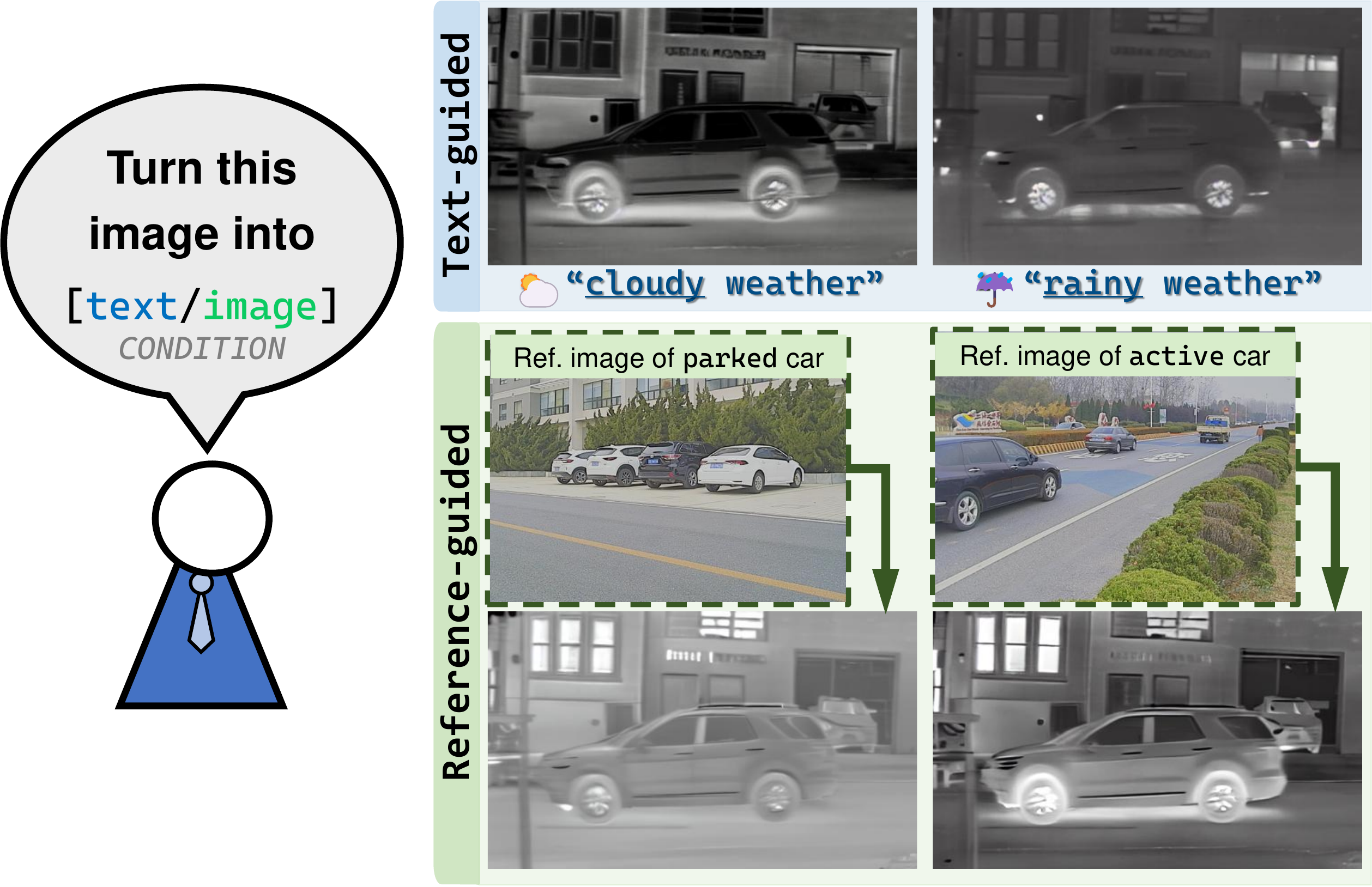}
    \caption{Controllable RGB-to-TIR translation}
    \label{controllable-fig1}
  \end{subfigure}
  \vspace{-7mm}
  \caption{
  \textbf{Overview of TherA.}
  (a) Compared to InstructPix2Pix \cite{brooks2023instructpix2pix}, our thermal-aware VLM distinguishes between active (heat-emitting car) and passive (parked car) objects as shown in the translated image from WayMo \cite{sun2020waymo} open dataset. 
  (b) TherA is the first RGB-to-\ac{TIR} translation model offering both text-guided and reference image-guided translations. RGB images from M3FD \cite{m3fd} are used for reference}
  \label{main_fig_overview}
  \vspace{-7mm}
\end{figure}

\Acf{TIR} imaging provides robust perception in low-visibility conditions, making it valuable both as a standalone sensor and as a complement to RGB imagery for diverse vision tasks \cite{thermalchameleon, shin2024complementary, vtmot, kniaz2018thermalgan}. However, building large annotated thermal datasets remains costly and slow due to expensive sensors and the expert effort required to label low-texture imagery. Physics-based \ac{TIR} simulators partly mitigate data scarcity but typically cover limited scenes and require substantial engineering to encode object-specific thermal behavior \cite{madan2023thermalsynth}. Consequently, many pipelines generate annotated pseudo-\ac{TIR} images via RGB-to-\ac{TIR} translation \cite{matchanything, minima, eccv_binding, imagebind, languagebind}.

The core challenge of RGB-to-TIR translation is that \ac{TIR} appearance is not a simple restyling of RGB. Most \ac{TIR} cameras and public datasets depict a scene’s relative temperature; thus, at a high level, \ac{TIR} appearance depends on temperature-dependent factors such as material (emissivity), active heat sources, and environments (time of day, season, weather) \cite{vollmer2018infrared}. This variability means a single RGB input can correspond to many valid thermal outputs, making the RGB-to-\ac{TIR} mapping inherently ambiguous.

Despite this ambiguity, many current RGB-to-\ac{TIR} models implicitly cast the translation task as pixel-level style transfer. This yields an ill-posed problem in which predicting thermal intensity solely from RGB pixels, without conditioning on thermally relevant attributes, often produces artificial outputs that contradict plausible \ac{TIR} physics. For example, in \cref{thermal-aware-fig1}, InstructPix2Pix \cite{brooks2023instructpix2pix} translates both cars with hot exhausts and wheels even though one is parked, revealing a clear inconsistency with thermal physics.
Recent translation methods attempt to address this limitation by incorporating segmentation maps \cite{ran2025diffv2ir, paranjape2025fvita} or scene-index priors \cite{xiao2025thermalgen} as auxiliary conditions to guide translation and alleviate the ill-posedness.
However, these cues encode category and layout information rather than the physics of heat emission and transfer, leaving the underlying problem unresolved.

In this paper, we present \textbf{TherA}, an RGB-to-\ac{TIR} translation framework conditioned by a \underline{Ther}mal-\underline{A}ware vision-language model (TherA-VLM). As shown in \cref{thermal-aware-fig1}, TherA produces thermally plausible outputs consistent with the thermal physics: the active car exhibits elevated thermal emissions while the parked car remains comparatively cool, and the building displays a realistic, spatially coherent heat distribution.

To support thermally grounded RGB-to-\ac{TIR} translation, we first construct \textbf{R2T2}, a 100k-pair dataset containing aligned RGB-\ac{TIR} images and structured thermal descriptions. These descriptions are produced by multimodal LLMs that jointly analyze RGB-\ac{TIR} inputs and characterize how thermally-relevant factors such as scene (e.g., time-of-day, weather, locations), materials, and object activity manifest in the thermal domain. R2T2 provides TherA-VLM with consistent supervisory signals for learning to infer thermally relevant attributes from RGB alone.

Using R2T2, we train \textbf{TherA-VLM}, a thermal-aware vision-language model that predicts a compact thermal embedding from an input RGB image and a user instruction. This thermal embedding conveys the thermal context of the scene and is used to condition our diffusion-based translator, guiding it to generate physically plausible \ac{TIR} outputs. Through this mechanism, TherA mitigates the ill-posedness of prior RGB-to-\ac{TIR} methods and produces consistent, physically grounded results.

Beyond improving translation quality, TherA-VLM also enables controllability: the user can modulate thermal appearance using either natural-language prompts or a reference RGB image, while the scene geometry remains unchanged. To the best of our knowledge, TherA is the first RGB-to-\ac{TIR} translation framework that provides such interpretable and physically meaningful control.

In summary, our main contributions are:
\begin{itemize}
\item \textbf{Thermal-aware VLM conditioning.}
As a key component of TherA, we propose TherA-VLM, which predicts a compact thermal embedding used to guide a diffusion-based RGB-to-\ac{TIR} translator.

\item \textbf{Controllable thermal modulation.}
TherA supports both text-guided and reference image-guided control over thermal appearance without altering scene geometry.

\item \textbf{State-of-the-art performance.}
Our method achieves \ac{SOTA} results on FLIR~\cite{fliradas} and M3FD~\cite{m3fd}, and exhibits strong zero-shot generalization.

\item \textbf{Dataset and model release.}
We will publicly release the R2T2 dataset, TherA-VLM weights, and translation module to facilitate research in \ac{TIR} imaging.
\end{itemize}

\vspace{0.3em}
\noindent

\section{Related Works}
\label{sec:related}

\subsection{RGB-to-Thermal Image Translation}

Recent efforts in RGB-to-\ac{TIR} translation have explored various priors to bridge the spectral gap. Many large-scale frameworks generate pseudo-\ac{TIR} data for pretraining \cite{minima,matchanything,jiang2024infrared, languagebind}, but they typically reuse existing style-transfer-based methods, inheriting their limitations in thermal plausibility. To move beyond simple style transfer, subsequent works introduced auxiliary RGB-centric priors. These range from low-level cues like edges \cite{ lee2023edge_srgb-tir, infragan, irformer} and segmentation maps \cite{ran2025diffv2ir, paranjape2025fvita} to scene-level cues like dataset indices \cite{xiao2025thermalgen}. While these priors improve geometric alignment, they remain blind to the factors that truly govern thermal emission: material emissivity, object state, and environmental contexts such as weather \cite{vollmer2018infrared}. PID \cite{pid-tevnet} use physics-based losses, but rely on a dataset-specific feature extractor, which limits generalization to unseen data.

Crucially, none of these methods provide a mechanism for explicit control. They enforce a deterministic one-to-one mapping, collapsing the inherently one-to-many nature of thermal imagery and limiting the diversity of pseudo-\ac{TIR} data. In contrast, our approach introduces a thermal-aware vision-language prior that grounds diffusion in semantic-physical reasoning, enabling generalizable and controllable RGB-to-\ac{TIR} translation.

\subsection{Controllable Image Translation}

Instruction-guided diffusion models have advanced controllable image editing by coupling textual and visual inputs.
InstructPix2Pix \cite{brooks2023instructpix2pix} first trained conditional diffusion models on synthetic instruction-image pairs, establishing the paradigm of instruction-based editing.
Adapter-based extensions \cite{ye2023ipadaptor,controlnet} inject external conditions into frozen Stable Diffusion backbones for stylization and layout control, yet these modules remain largely style-transfer oriented and are limited to RGB domains.
MGIE \cite{fu2023guiding}, SmartEdit \cite{huang2024smartedit}, and GoT \cite{fang2025got} extended this idea to multimodal reasoning through chain-of-thought parsing of complex instructions.
Despite their controllability, these systems still rely on verbose, free-form prompts and lack grounding in physical modalities.
Our work draws inspiration from this progression, aiming to extend vision-language conditioning beyond RGB semantics toward physically meaningful, thermally grounded translation.

\section{Methods}
\label{sec:methods}

\begin{figure*}[t]
 \centering
 \includegraphics[width=\textwidth]{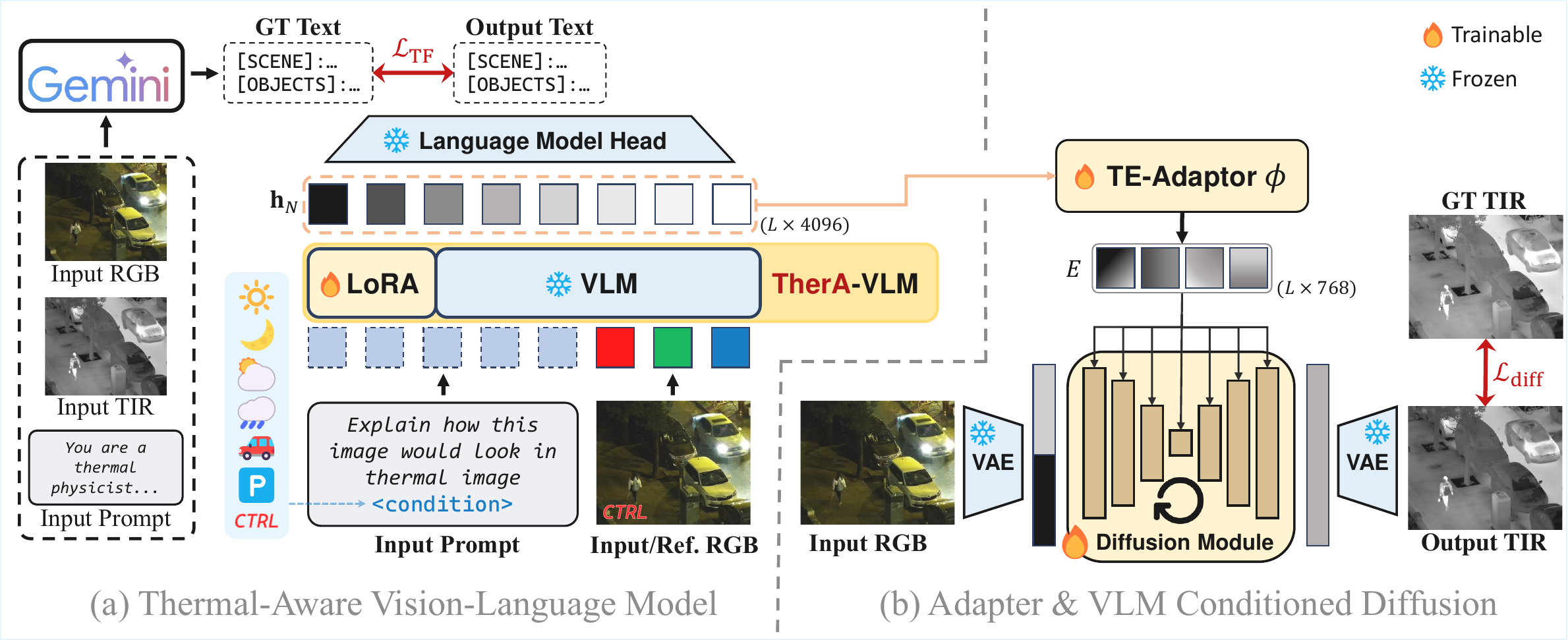} 
 \vspace{-8mm}
 \caption{
  \textbf{Overview of the TherA framework.}
  Our model consists of two stages.
  \textbf{(a) TherA-VLM} analyzes the input/reference RGB and the input prompt to produce a physically grounded thermal embedding ($\textbf{h}_N$).
  \textbf{(b)} This thermal embedding is injected via \textbf{TE Adapter} ($\phi$) into the cross-attention layers of a UNet.
  The UNet, also conditioned on the RGB latent ($z_{\mathrm{rgb}}$), guides the denoising of the noisy TIR latent ($z_t$) to generate a physically plausible TIR image. The input prompt and reference RGB supports controllability.
 }
 \label{fig:framework_pipeline}
 \vspace{-4mm}
\end{figure*}

An overview of the TherA framework is illustrated in \cref{fig:framework_pipeline}.
It comprises two synergistic components that together enable semantically grounded and physically consistent RGB-to-TIR translation.
First, illustrated in \figref{fig:framework_pipeline}\textcolor{cvprblue}{a}, TherA-VLM interprets the input RGB image, reasoning about its physical properties to generate structured scene descriptors that encode objects, materials, and their heat-emission states.
Second, illustrated in \figref{fig:framework_pipeline}\textcolor{cvprblue}{b} we introduce a VLM-conditioned denoising diffusion model where the UNet is conditioned on the output thermal embedding ($\textbf{h}_N$) of TherA-VLM.
This design replaces the conventional CLIP text embedding, enabling the diffusion process to generate realistic TIR imagery that aligns with both physical priors and scene context.
Together, these two modules establish a deterministic and  controllable generation pipeline.

\subsection{R2T2 Dataset}
We construct an RGB-TIR-text dataset, R2T2, to train both TherA-VLM and the diffusion model. R2T2 contains 100k triplets, each consisting of an RGB image, its aligned TIR counterpart, and a canonical text describing the thermal characteristics of the scene and objects (class, material, color) and their heat emission states.

To build R2T2, we first employ a multimodal reasoning model (Gemini 2.5 Pro \cite{team2024gemini}) that jointly analyzes paired RGB-TIR inputs to describe how the RGB scene manifests in the thermal domain, producing keyword-oriented structured outputs. We then canonicalize these reasoning-based captions by merging synonyms and standardizing object attributes (see \cref{sec:R2T2_curation}).
This canonicalization regularizes the conditioning space and removes linguistic variability, yielding consistent supervision for TherA-VLM.

R2T2 is compiled from nine existing aligned datasets~\cite{vtmot,xoftr-metu-vistir,kaist-mdpd,llvip,sjtu,aviid,msrs,fliradas,m3fd}, and to further increase scene diversity, we additionally curate pseudo-aligned RGB-TIR pairs from three synchronous but originally unaligned datasets~\cite{camel,vivid++,nsavp}. This provides broad coverage across viewpoints, seasons, locations, and environmental conditions.

We reserve M3FD~\cite{m3fd} and FLIR~\cite{fliradas} in R2T2 for validation and use all remaining data to train both TherA-VLM and the diffusion model. Additional details of the curation pipeline are provided in \cref{sec:R2T2_curation}.

\subsection{Thermal-Aware Vision--Language Model}
\label{sec:ta_vlm}

\begin{figure}[t]
    \centering
    \begin{subfigure}[t]{0.49\linewidth}
        \centering
        \includegraphics[width=\linewidth]{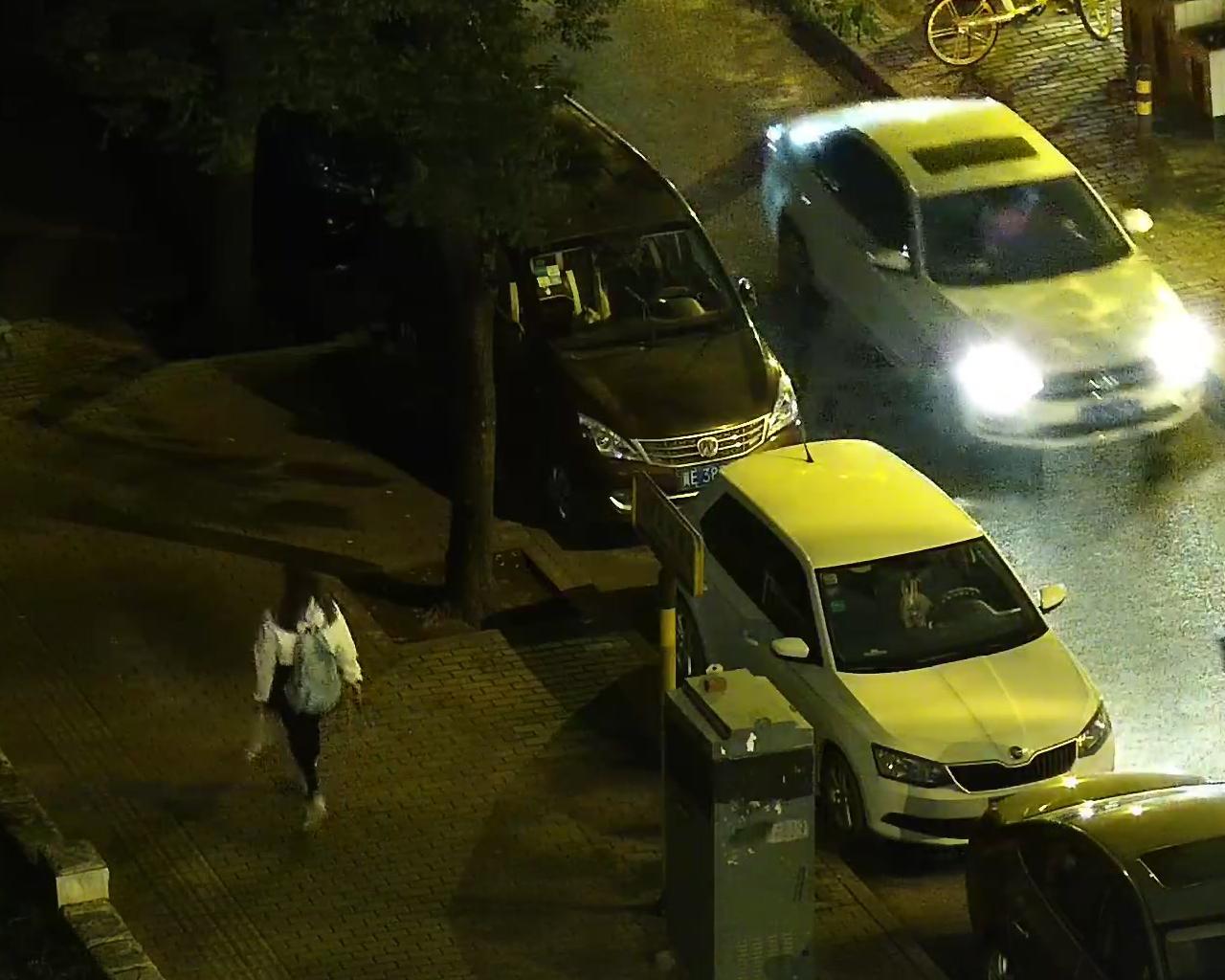} 
        \caption*{\textbf{(a)} RGB input}
    \end{subfigure}
    \hfill
    \begin{subfigure}[t]{0.49\linewidth}
        \centering
        \includegraphics[width=\linewidth]{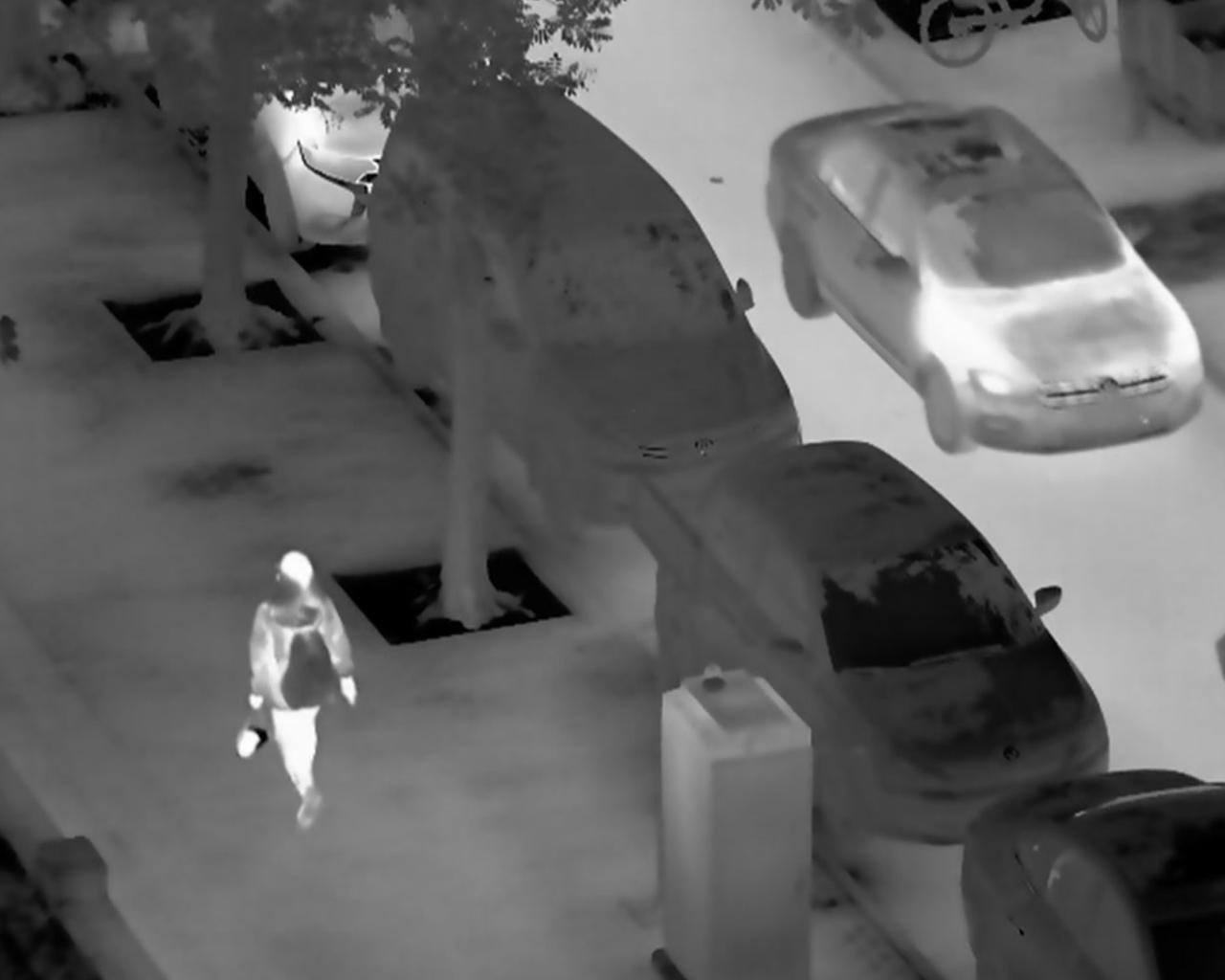} 
        \caption*{\textbf{(b)} Ground-truth TIR}
    \end{subfigure}
    
    \vspace{3pt}
    
    \begin{subfigure}[t]{0.49\linewidth}
        \centering
        \includegraphics[width=\linewidth]{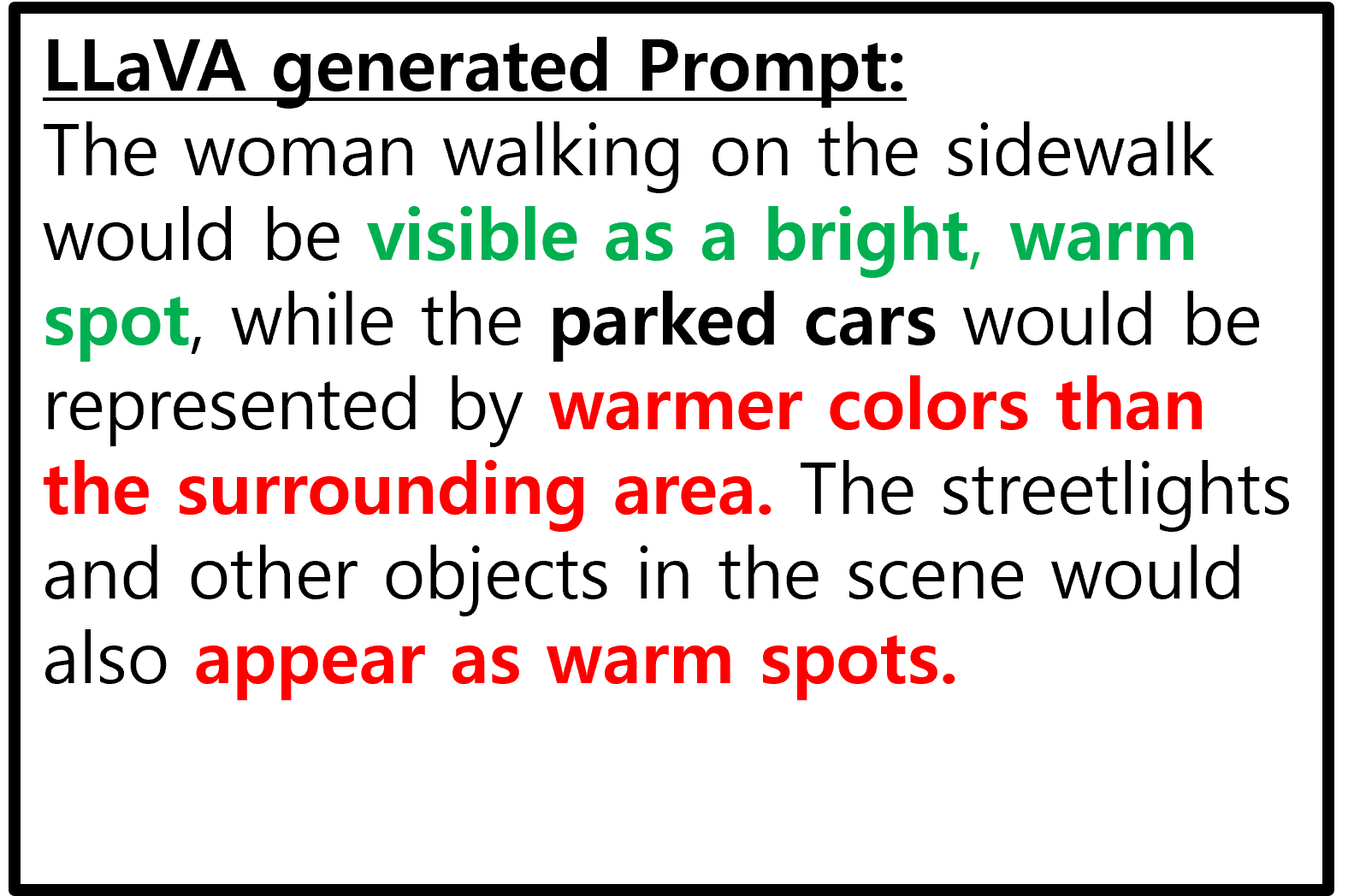} 
        \caption*{\centering\textbf{(c)} LLaVA\\ (RGB-centric caption)}
        \label{llava-prose}
        
    \end{subfigure}
    \hfill
    \begin{subfigure}[t]{0.49\linewidth}
        \centering
        \includegraphics[width=\linewidth]{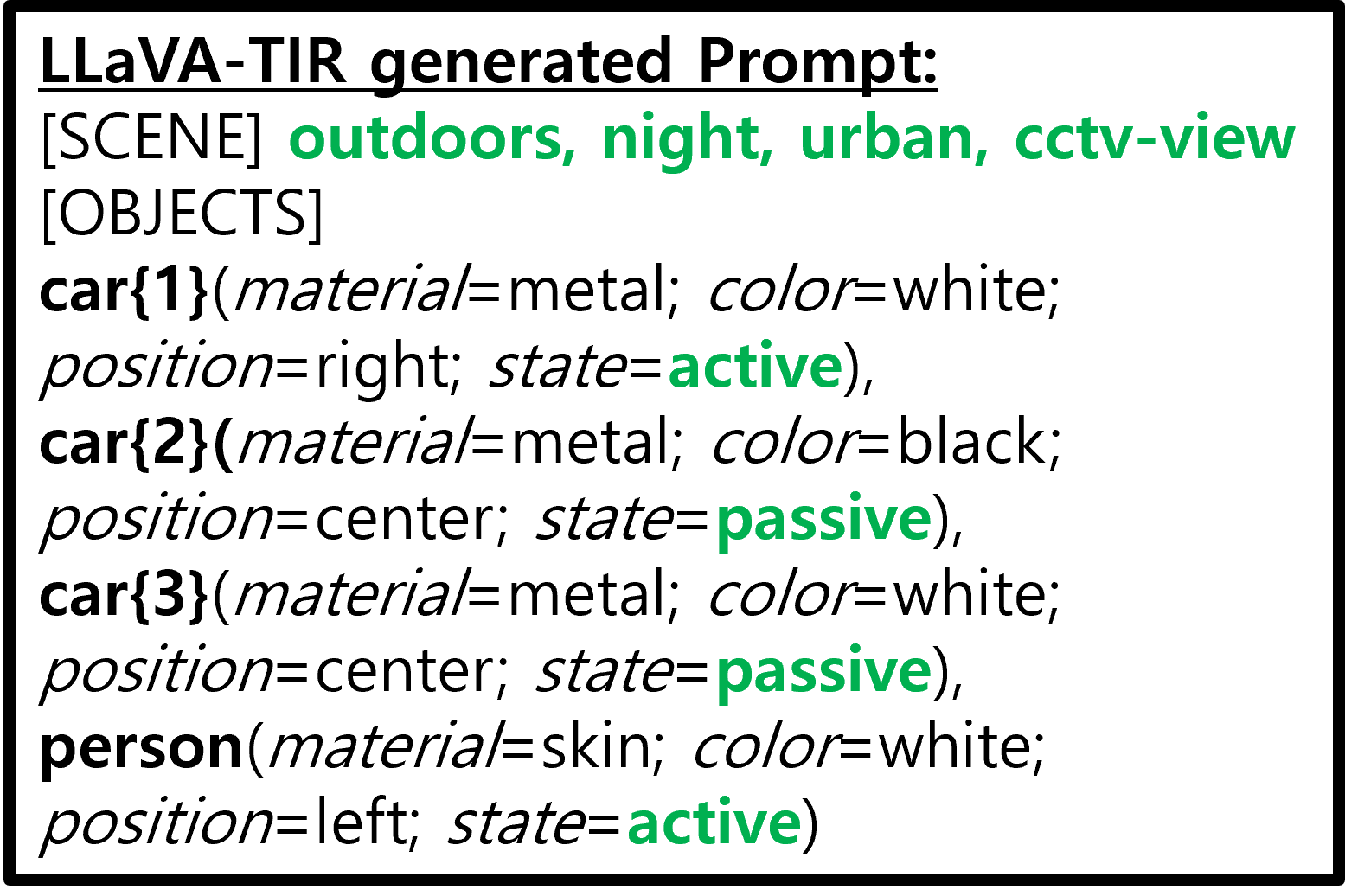} 
        \caption*{\centering\textbf{(d)} TherA-VLM\\ (Thermal-aware canonical schema)}
        \label{our-prompt}
        
    \end{subfigure}
    
    \vspace{1.5pt}
    
    \caption{
        Comparison of textual conditioning.
        Standard LLaVA produces appearance-based, RGB-centric descriptions (c).
        Our \textbf{TherA-VLM} outputs concise, structured schemas encoding scene type, materials, and heat emissionstates (d).
    }
    \label{fig:thermalvlm_compare}
    \vspace{-4mm}
\end{figure}

A key challenge in language-guided RGB-to-\ac{TIR} translation lies in defining physically meaningful textual conditioning.
Most existing methods~\cite{ran2025diffv2ir, paranjape2025fvita, minima} treat the language branch merely as a stylistic cue, using generic prompts such as ``turn this into a thermal infrared image'' or short captions generated by BLIP~\cite{li2022blip}.
Such RGB-centric descriptions (\cref{fig:thermalvlm_compare}) often result in vague conditioning and poor thermal plausibility.

To enable physically grounded conditioning, we designed TherA-VLM, denoted $f_{\theta}$, to output structured and canonicalized schemas summarizing thermally relevant attributes.
Given an RGB input $I_{\mathrm{rgb}}$ and user prompt $p_{\mathrm{user}}$, the model produces
\begin{equation}
S = f_{\theta}(I_{\mathrm{rgb}},p_{\mathrm{user}}) = 
\{ s_{\mathrm{scene}},\, s_{\mathrm{object}},\, s_{\mathrm{material}},\, s_{\mathrm{heat}} \}
\end{equation}
where each element encodes the scene, objects, materials, and heat emission states.
Unlike free-form captions, this schema provides a compact, interpretable representation suitable for multi-modal grounding.
As illustrated in \cref{fig:thermalvlm_compare}, conventional VLMs (\figref{fig:thermalvlm_compare}\textcolor{cvprblue}{c}) produce verbose and unstructured RGB-centric captions that describe \textit{what} appears in the scene but fail to capture \textit{how} it radiates heat. In contrast, our TherA-VLM (\figref{fig:thermalvlm_compare}\textcolor{cvprblue}{d}) generates concise, structured schemas detailing these physically relevant attributes.

\paragraph{Fine-tuning and Output Representation.}
We use LLaVA~1.5~\cite{liu2024improvedllava} as the base model for our TherA-VLM, fine-tuning only the LoRA layers of its language and image-projection modules on the canonical schema pseudo-labels via teacher forcing.
Each training sample consists of an RGB input $I_{\mathrm{rgb}}$ and schema sequence $y=\{y_n\}_{n=1}^{N}$, optimized via teacher-forcing loss
\begin{equation}
\mathcal{L}_{\mathrm{TF}} = -\sum_{n=1}^{N}\log p_{\theta}(y_n \mid y_{<n}, I_{\mathrm{rgb}}),
\end{equation}
which encourages the model to autoregressively predict thermally grounded tokens from visual cues.
During inference, $f_{\theta}$ produces the canonical schema from the input RGB image, and its final hidden state, denoted as thermal embedding $\textbf{h}_N$ serves as the compact semantic embedding used to condition the diffusion UNet.

\subsection{VLM-Conditioned Diffusion}
\label{sec:vlm_diffusion}

\paragraph{Network Architecture}
We employ a UNet-based latent diffusion model \cite{stablediffusion}. The UNet's input is modified to accept 8 channels: 4 channels for the noisy target \ac{TIR} latent $z_t$, and 4 channels for the RGB guidance latent $z_{\mathrm{rgb}}$.
Let $I_{\mathrm{rgb}}, I_{\mathrm{tir}} \in [-1,1]^{H \times W \times 3}$ be the input images.
A frozen VAE encoder $\mathcal{E}$ from the Stable Diffusion model~\cite{stablediffusion} maps images to latents $\mathcal{Z} \subset \mathbb{R}^{h \times w \times 4}$ with scaling factor $\alpha$:
\begin{equation}
z_{\mathrm{rgb}} = \alpha \cdot \mathcal{E}(I_{\mathrm{rgb}}), 
\qquad
z_{\mathrm{tir}} = \alpha \cdot \mathcal{E}(I_{\mathrm{tir}}).
\end{equation}
During training, Gaussian noise $\epsilon \sim \mathcal{N}(0, \mathbf{I})$ is added at timestep $t$ to $z_{\mathrm{tir}}$, yielding $z_t$. The full 8-channel UNet input at timestep $t$ is:
\begin{equation}
\mathbf{x}_t = \mathrm{concat}\big(z_t,\, z_{\mathrm{rgb}}\big) \in \mathbb{R}^{h \times w \times 8}.
\end{equation}

\paragraph{Bridging VLM embeddings to UNet.}
Directly replacing CLIP embedding with hidden states from a large VLM is nontrivial, as the UNet must adapt to both a new embedding space and a new \ac{TIR} output domain.
To address this, we take inspiration from LaVi-Bridge~\cite{zhao2024bridging}, which stabilizes alignment through a separate adapter module.
Specifically, since our LLaVA-based TherA-VLM and the LLaMa--Stable Diffusion UNet from \cite{zhao2024bridging} share a LLaMA lineage, we employ the adapter module \cite{zhao2024bridging} $\phi$ that projects the TherA-VLM thermal embedding $\textbf{h}_N \in \mathbb{R}^{L \times 4096}$ into the UNet attention width:
\begin{equation}
E = \phi(\textbf{h}_N) \in \mathbb{R}^{L \times 768}.
\end{equation}
This thermal embedding (TE) adapter is a two-layer feedforward network that bridges the VLM semantics with the UNet cross-attention layers. During the diffusion training, we freeze TherA-VLM and optimize only the UNet and the TE adapter.
This design effectively transfers cross-modal alignment from large-scale pretraining to our RGB-to-\ac{TIR} task.
The training objective $\mathcal{L}_{\mathrm{diff}}$ for the UNet $\hat{\epsilon}$ is:
\begin{equation}
\mathcal{L}_{\mathrm{diff}} = \mathbb{E}_{t, \epsilon, E, M}\left[ \left\| \hat{\epsilon}(\mathbf{x}_t, t, E, M) - \epsilon \right\|_2^2 \right],
\end{equation}
where $E$ is the projected thermal embedding and $M$ is the VLM's attention mask.

\paragraph{Dual-CFG Inference.}
Like previous works \cite{brooks2023instructpix2pix}, we employ dual \ac{CFG} on both the image ($c_V$) and VLM ($c_S$) conditions. During inference, the guided noise estimate $\tilde{\epsilon}_{\theta}$ is computed progressively:
\begin{equation}
\begin{aligned}
\tilde{\epsilon}_{\theta}(\mathbf{x}_t, c_V, c_S)
&= \epsilon_{\theta}(\mathbf{x}_t, \varnothing, \varnothing) \quad \text{(unconditional)} \\
&\quad + s_V \!\cdot\! \big(\epsilon_{\theta}(\mathbf{x}_t, c_V, \varnothing) - \epsilon_{\theta}(\mathbf{x}_t, \varnothing, \varnothing)\big) \\
&\quad + s_S \!\cdot\! \big(\epsilon_{\theta}(\mathbf{x}_t, c_V, c_S) - \epsilon_{\theta}(\mathbf{x}_t, c_V, \varnothing)\big),
\end{aligned}
\label{eq:dual_cfg_simple}
\end{equation}
where $s_V$ and $s_S$ are the guidance scales. This formulation first incorporates visual structure ($s_V$) and then layers thermally grounded semantics ($s_S$).

\paragraph{Controllability.}
A key benefit of our design is that controllability emerges naturally from manipulating the VLM's inputs (text-prompt, reference image), with the VLM's thermal embedding $\textbf{h}_N$ adapting accordingly to impose targeted modifications while ensuring thermally plausible outputs. TherA supports two levels of control:

\textbf{(i) Text-guided Control:} Text-guided control can be achieved by concatenating user instruction after input prompt $p_{\mathrm{user}}$. This updates the VLM's schema $S$, generating a new global thermal embedding $\textbf{h}_N$ that influences the output. Currently, we found that text-guided control can be used only for controlling scene-wise attributes (e.g., cloudy weather, nighttime).


\textbf{(ii) Reference-guided Control:} Our method also supports reference-guided control. In a typical inference pass, the same RGB image is inputted to both TherA-VLM and the diffusion module. For guided synthesis, however, a different reference RGB image which possesses the desired attributes is provided to the TherA-VLM. This prompts TherA-VLM to extract a thermal embedding specific to this reference. This thermal embedding $E=\phi(\textbf{h}_N)$ is then used to condition the diffusion module, directly influencing the generated TIR output. We found that this provides effective control over not only global, scene-wise attributes but also granular, object-wise properties (e.g., changing the thermal state of a specific car). 


\section{Experiments}
\label{sec:experiments}

\begin{table*}[t]
\centering
\caption{RGB-to-TIR image translation evaluation on M3FD and FLIR. Best results are highlighted in \textbf{bold} and second best are \underline{underlined}.}
\vspace{-3mm}
\resizebox{0.7\textwidth}{!}{%
\begin{tabular}{cl|cccc|cccc}
\toprule
\multicolumn{2}{c|}{\multirow{2}{*}{\textbf{Method}}} &
\multicolumn{4}{c|}{\textbf{M3FD} \cite{m3fd}} &
\multicolumn{4}{c}{\textbf{FLIR} \cite{fliradas}} \\
\cmidrule(lr){3-6} \cmidrule(lr){7-10}
 & \textbf{} &
PSNR$\uparrow$ & SSIM$\uparrow$ & FID$\downarrow$ & LPIPS$\downarrow$ &
PSNR$\uparrow$ & SSIM$\uparrow$ & FID$\downarrow$ & LPIPS$\downarrow$ \\
\midrule

\multirow{8}{*}{\rotatebox{90}{\textbf{General I2I}}} 
& InstructPix2Pix \cite{brooks2023instructpix2pix} & 13.94 & 0.40 & 138.94 & 0.37 & 14.44 & 0.38 & 178.03 & 0.41 \\
& StegoGAN \cite{wu2024stegogan} & 13.14 & 0.30 & 247.03 & 0.47 & 12.58 & 0.29 & 152.57 & 0.46 \\
& Pix2Pix-Turbo \cite{parmar2024pix2pixturbo} & 16.69 & 0.60 & 142.26 & 0.32 & 17.03 & 0.40 & 91.17 & \underline{0.32} \\
& UNSB \cite{kim2024unsb} & 12.86 & 0.39 & 160.72 & 0.47 & 16.91 & 0.42 & 139.46 & 0.40 \\
& CSGO \cite{xing2024csgo} & 9.61 & 0.36 & 204.53 & 0.56 & 10.12 & 0.28 & 116.85 & 0.55 \\ 
& StyleID \cite{chung2024styleid} & 12.64 & 0.44 & 155.69 & 0.43 & 12.66 & 0.28 & 164.43 & 0.39 \\  
& StyleSSP \cite{xu2025stylessp} & 13.28 & 0.51 & 165.92 & 0.50 & 10.40 & 0.35 & 150.52 & 0.50 \\ 
& BBDM \cite{li2023bbdm} & 16.43 & 0.57 & 238.60 & 0.45 & 16.10 & 0.45 & 177.81 & 0.42 \\ 
\midrule

\multirow{8}{*}{\rotatebox{90}{\textbf{RGB-to-TIR I2I}}} 
& InfraGAN \cite{infragan} & 14.49 & 0.55 & 254.55 & 0.39 & 16.97 & 0.40 & 224.00 & 0.37 \\
& IR-Former \cite{irformer} & 17.88 & 0.58 & 196.80 & 0.43 & 17.64 & 0.48 & 214.22 & 0.50 \\ 
& EG-GAN \cite{lee2023edge_srgb-tir} & 12.99 & 0.45 & 163.78 & 0.45 & 14.30 & 0.37 & 140.92 & 0.55 \\ 
& DR-AVIT \cite{dr-avit} & 14.63 & 0.52 & 161.32 & 0.39 & 15.11 & 0.35 & 111.01 & 0.39 \\
& PID \cite{pid-tevnet} & 16.10 & 0.56 & 160.91 & 0.46 & 17.26 & 0.40 & \underline{84.26} & 0.36 \\ 
& F-ViTA \cite{paranjape2025fvita} & 17.36 & 0.61 & 111.77 & 0.29 & 17.76 & 0.43 & 87.03 & 0.33 \\
& DiffV2IR \cite{ran2025diffv2ir} & \underline{18.97} & \underline{0.66} & \underline{92.57} & \underline{0.23} & \underline{18.24} & \underline{0.48} & 91.44 & 0.33 \\
& ThermalGen \cite{xiao2025thermalgen} & 15.90 & 0.51 & 110.74 & 0.43 & 17.59 & 0.47 & 101.45 & 0.41 \\
\midrule
& \textbf{TherA} & \textbf{19.54} & \textbf{0.67} & \textbf{87.08} & \textbf{0.21} & \textbf{19.02} & \textbf{0.53} & \textbf{83.78} & \textbf{0.31} \\ 
\bottomrule
\end{tabular}%
}
\label{translation_results_benchmark}
\vspace{-2mm}
\end{table*}
\begin{table*}
\centering
\caption{Zero-shot evaluation of RGB-to-TIR translation on M3FD, FLIR and CART.}
\vspace{-3mm}
\resizebox{0.8\textwidth}{!}{%
\begin{tabular}{l|cccc|cccc|cccc}
\toprule
\multicolumn{1}{c|}{\multirow{2}{*}{\textbf{Method}}} &
\multicolumn{4}{c|}{\textbf{M3FD} \cite{m3fd}} &
\multicolumn{4}{c}{\textbf{FLIR} \cite{fliradas}} &
\multicolumn{4}{c}{\textbf{CART} \cite{lee2024caltech-cart}}\\
\cmidrule(lr){2-5} \cmidrule(lr){6-9} \cmidrule(lr){10-13}
 & 
PSNR$\uparrow$ & SSIM$\uparrow$ & FID$\downarrow$ & LPIPS$\downarrow$ &
PSNR$\uparrow$ & SSIM$\uparrow$ & FID$\downarrow$ & LPIPS$\downarrow$ &
PSNR$\uparrow$ & SSIM$\uparrow$ & FID$\downarrow$ & LPIPS$\downarrow$\\
\midrule
F-ViTA \cite{paranjape2025fvita} & \underline{13.86} & 0.46 & 145.83 & \underline{0.42} & 13.86 & \underline{0.45} & 145.83 & \underline{0.42} & 11.25 & 0.44 & 221.83 & 0.65\\
DiffV2IR \cite{ran2025diffv2ir} & 11.77 & \underline{0.49} & \underline{132.49} & 0.46 & 11.41 & 0.18 & 253.82 & 0.66 & 10.92 & 0.43 & \underline{184.93} & \underline{0.52}\\
ThermalGen \cite{xiao2025thermalgen} & 12.84 & 0.35 & 177.85 & 0.54 & \underline{14.06} & 0.39 & \underline{127.14} & 0.55 & \underline{12.17} & \underline{0.49} & 215.73 & 0.61 \\
\textbf{TherA} & \textbf{18.24} & \textbf{0.65} & \textbf{105.52} & \textbf{0.29} & \textbf{16.56} & \textbf{0.49} & \textbf{112.93} & \textbf{0.38} & \textbf{15.38} & \textbf{0.60} & \textbf{169.33} & \textbf{0.36} \\ 
\bottomrule
\end{tabular}%
}
\vspace{-5mm}
\label{translation_results_zeroshot}
\end{table*}


\subsection{Experimental Setup}
\label{sec:setup}

\paragraph{Datasets.}

We trained TherA-VLM with R2T2, and for image translation we used M3FD \cite{m3fd} and FLIR \cite{fliradas} for standard training and evaluation. For zero-shot experiments, models were trained on R2T2 and evaluated on M3FD, FLIR, and CART \cite{lee2024caltech-cart}.

\vspace{-5mm}
\paragraph{Evaluation Metrics.}
We evaluate translation quality using Peak Signal-to-Noise Ratio (PSNR), Structural Similarity Index (SSIM), Fréchet Inception Distance (FID), and Learned Perceptual Image Patch Similarity (LPIPS).
\vspace{-3mm}
\paragraph{Implementation Details.}
All models were trained on 4 NVIDIA A6000 (48GB) GPUs. For TherA-VLM, we finetuned the vision projector and language model of a pretrained LLaVa \cite{liu2024improvedllava} backbone using LoRA \cite{hu2022lora} (r=128, a=256). We used the instruction prompt ``\textit{Explain how this image would look in thermal image} \verb|<|\texttt{condition}\verb|>|'', where the \texttt{condition} token was randomly dropped (10\% probability) to prevent overfitting. For the diffusion model, we initialized the UNet with Stable Diffusion 1.4 \cite{stablediffusion} and the adapter with the LLaMa-UNet Bridge \cite{zhao2024bridging}. The model was trained for 100 epochs using AdamW with a learning rate of $1e-4$ and a batch size of 32 per GPU. Further details are provided in \Cref{sec:additional_implementation_details}.


\begin{figure*}[t]
    \centering
    \begin{subfigure}{0.9\textwidth}
        \centering
        \includegraphics[width=\textwidth]{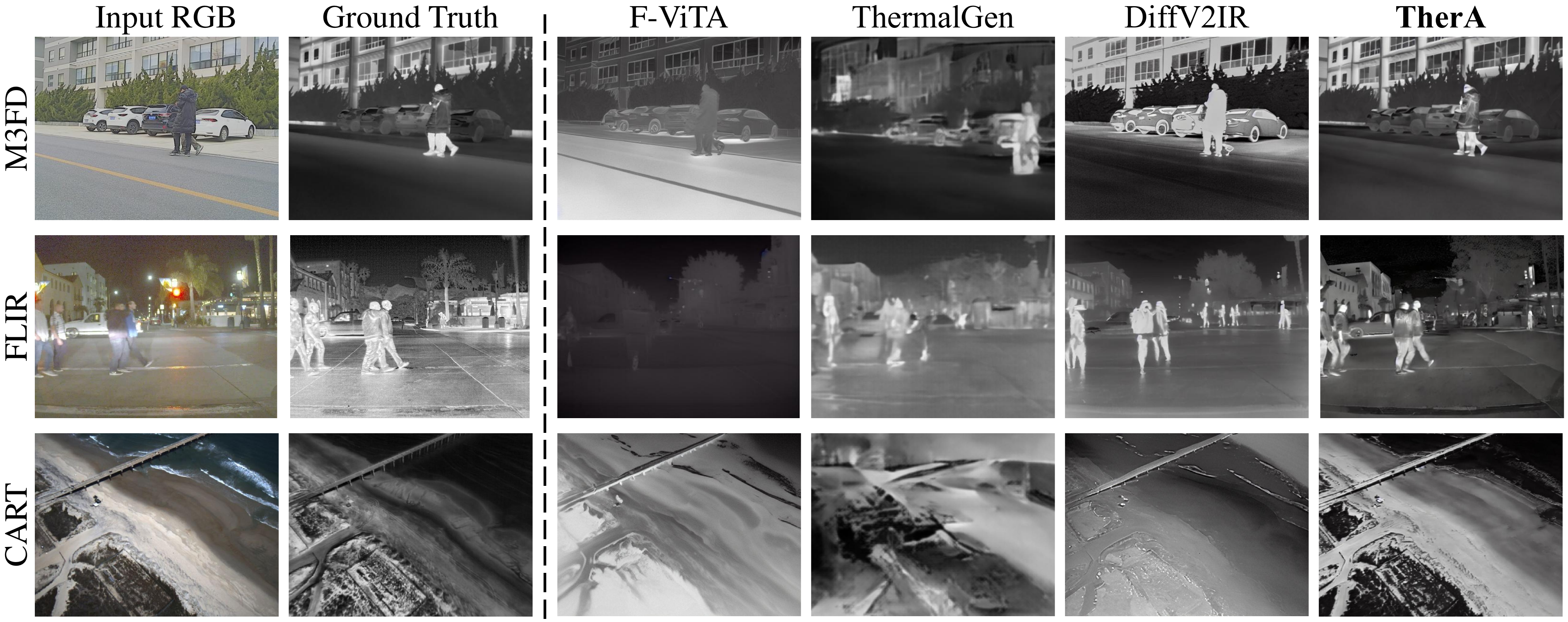}
        \caption{Zero-shot results for RGB-\ac{TIR} datasets: M3FD \cite{m3fd}, FLIR \cite{fliradas}, CART \cite{lee2024caltech-cart}.}
        \label{fig:rgb-t-qualitative}
    \end{subfigure}
    \vspace{1.5mm} 
    \begin{subfigure}{0.79\textwidth}
        \centering
        \includegraphics[width=\textwidth]{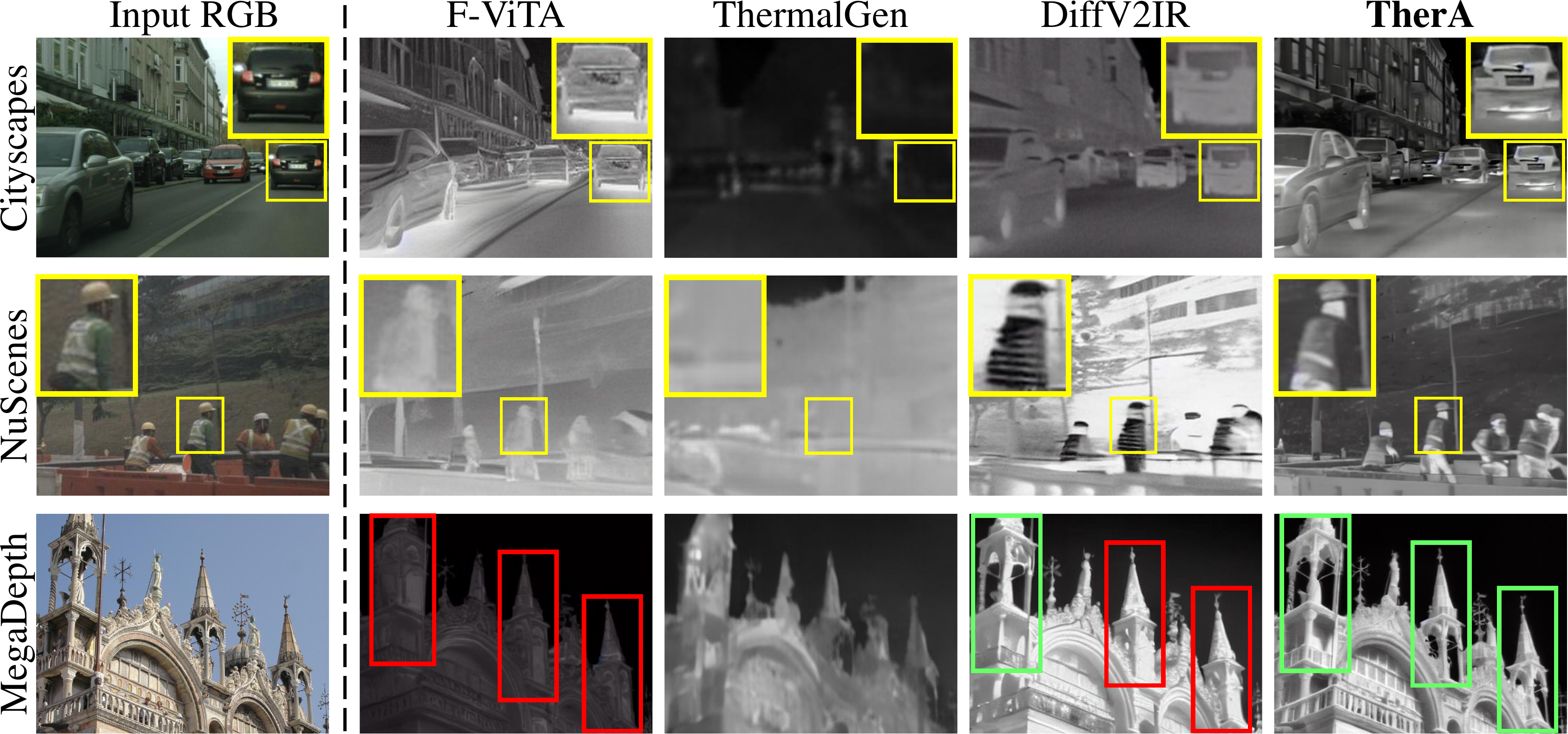}
        \caption{Zero-shot results for RGB-only datasets: Cityscapes \cite{cordts2016cityscapes}, NuScenes \cite{caesar2020nuscenes}, MegaDepth \cite{li2018megadepth}. \textcolor{orange}{Yellow} boxes show zoomed views, \textcolor{red}{Red} boxes show image artifacts from over-segmented images, \textcolor{green}{Green} boxes show correctly translated images.}
        \label{fig:rgb-qualitative}
    \end{subfigure}

    \vspace{-4mm}
    
    \caption{
        Qualitative results on zero-shot image translation. No \ac{TIR} ground truth exists for RGB-only datasets.
    }
    \label{fig:zeroshot-qualitative}
    \vspace{-4mm}
\end{figure*}

\subsection{Main Results on RGB-to-TIR Translation}

We evaluate TherA against a comprehensive suite of \ac{SOTA} baselines, including general \ac{I2I} models (paired \cite{brooks2023instructpix2pix, parmar2024pix2pixturbo, li2023bbdm} and unpaired \cite{kim2024unsb, wu2024stegogan}), example-guided \ac{I2I} models \cite{xing2024csgo, chung2024styleid, xu2025stylessp}, and specialized RGB-to-\ac{TIR} translators \cite{infragan,irformer,lee2023edge_srgb-tir,dr-avit,pid-tevnet,paranjape2025fvita,ran2025diffv2ir,xiao2025thermalgen}. For example guided-I2I models, we randomly sample a \ac{TIR} image from test set as the reference.

\begin{figure*}[t]
  \centering
  \begin{subfigure}{0.85\textwidth}
    \centering
    \includegraphics[width=\textwidth]{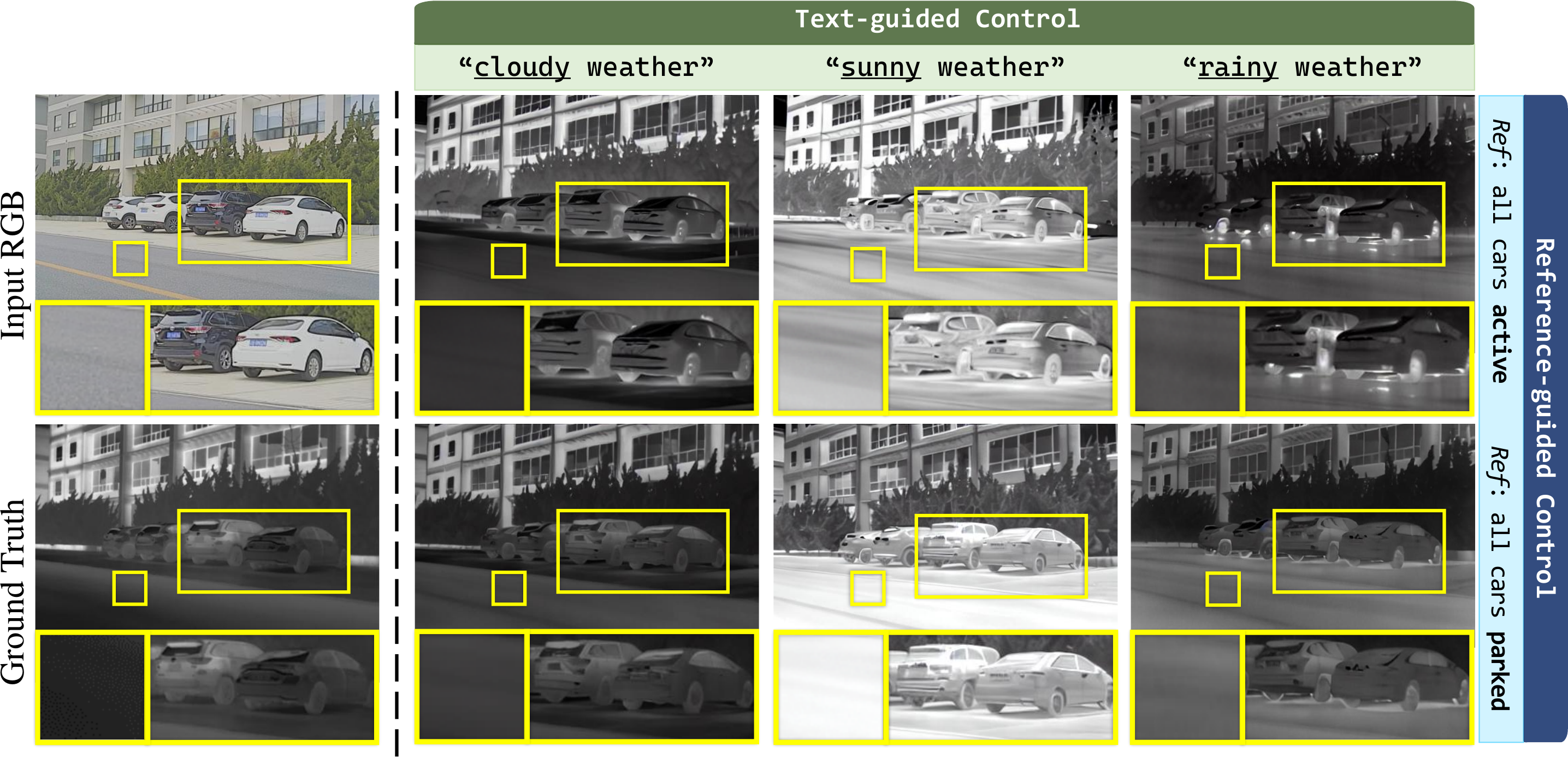}
    \caption{Text- and reference-guided controllability results of M3FD \cite{m3fd} on various weather and car states, respectively.}
    \label{controllability_results1}
  \end{subfigure}
  \begin{subfigure}{\textwidth}
    \centering
    \includegraphics[width=0.86\textwidth]{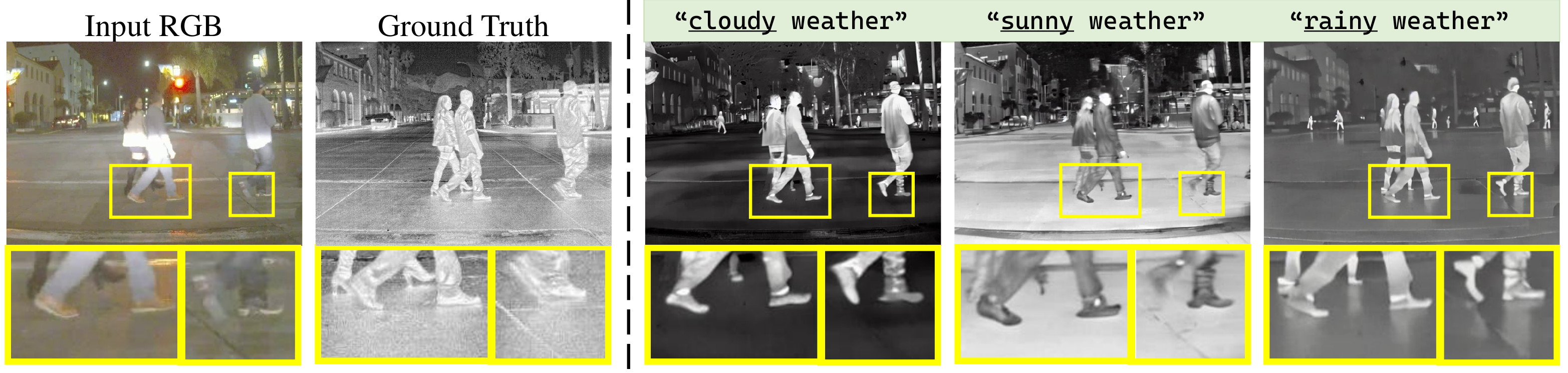}
    \caption{Text-guided controllability results of FLIR \cite{fliradas} on 'cloudy', 'sunny' and 'rainy' weather.}
    \label{controllability_results2}
  \end{subfigure}
  \begin{subfigure}{\textwidth}
    \centering
    \includegraphics[width=0.86\textwidth]{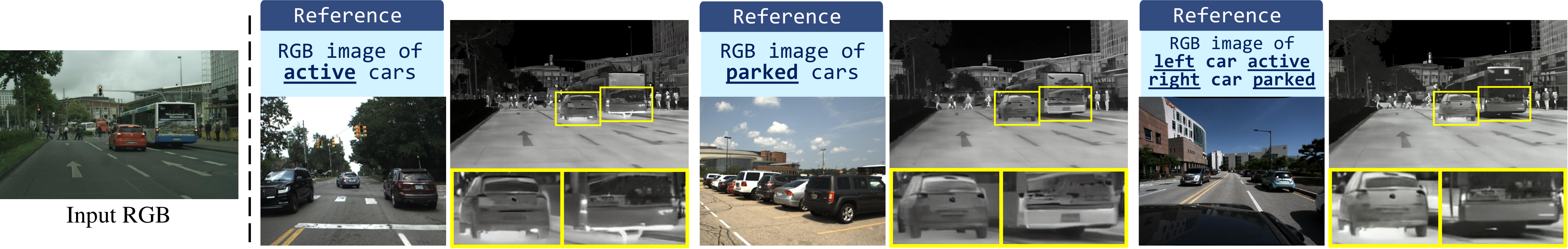}
    \caption{Reference-guided controllability results of Cityscapes \cite{cordts2016cityscapes} on reference RGB images depicting active, parked and mixed car states.}
    \label{controllability_results3}
  \end{subfigure}
  \caption{Qualitative results for scene-level and object-level controllability on M3FD and FLIR. \textcolor{orange}{Yellow} boxes show zoomed views.}
  \label{controllability_results}
  \vspace{-4mm}
\end{figure*}

\subsubsection{Quantitative Evaluation}


\paragraph{Benchmark Evaluation}
As shown in \cref{translation_results_benchmark}, we evaluate TherA against all \ac{SOTA} baselines on the M3FD \cite{m3fd} and FLIR \cite{fliradas} benchmarks, under the standard fine-tuning setting. All baselines are trained using their publicly available implementations and recommended hyperparameters.

TherA consistently outperforms all baseline models across all metrics and benchmarks. Among the baselines, translation models that incorporate additional prior conditions tend to perform better. For example, DiffV2IR \cite{ran2025diffv2ir} leverages both segmentation maps and scene-description priors, achieving the strongest results among prior methods. This observation supports our hypothesis that direct RGB-to-\ac{TIR} translation through purely end-to-end pixel learning leads to limited generalization, even within the same dataset, and that injecting structured priors can substantially improve physical consistency and translation quality.


\paragraph{Zero-shot Generalization.}
We further evaluate zero-shot generalization in \cref{translation_results_zeroshot} by testing the most recent \ac{SOTA} models (F-VITA \cite{paranjape2025fvita}, DiffV2IR \cite{ran2025diffv2ir}, and ThermalGen \cite{xiao2025thermalgen}) trained exclusively on R2T2, without benchmark-specific fine-tuning.

In this setting, TherA achieves the best zero-shot translation performance by a large margin across three different benchmarks, demonstrating strong generalization to unseen domains. We attribute this to the TherA-VLM, whose conditioning mechanism leverages the prior knowledge of a pretrained VLM to provide robust visual grounding from the input RGB image. By constraining the conditioning outputs to structured, keyword-oriented embeddings rather than free-form text, the model reduces linguistic noise and stabilizes training. This structured conditioning enables more reliable thermal reasoning, explaining the superior zero-shot performance compared to other models.

\subsubsection{Qualitative Evaluation}

\cref{fig:zeroshot-qualitative} provides visual comparisons of zero-shot translation results across both RGB-T paired and RGB-only datasets. ThermalGen’s scene-index priors capture global thermal tone but frequently miss fine-grained spatial structure, while segmentation-driven methods \cite{ran2025diffv2ir, paranjape2025fvita} preserve geometry yet struggle to reproduce distinctive \ac{TIR} appearance, often producing overly smooth or textureless heat patterns. These limitations become even more pronounced on RGB-only datasets, where over-segmentation from SAM \cite{kirillov2023SAM} illustrated in the last row of \cref{fig:zeroshot-qualitative}\textcolor{cvprblue}{b} results in structural distortions, while scene-index priors fail to generalize, leading to dark or artifact-heavy outputs. Additional results are available in \Cref{sec:zeroshot_qualitative}.


In contrast, TherA's thermal-aware conditioning preserves both thermal realism, by correctly identifying semantic-thermal properties (rendering active objects hot and inactive ones cool), and image fidelity, by maintaining complex structures without relying on explicit geometric priors such as segmentation maps.

\subsection{Controllable RGB-to-TIR Translation}
\begin{figure}[t]
 \centering
 \includegraphics[width=\columnwidth]{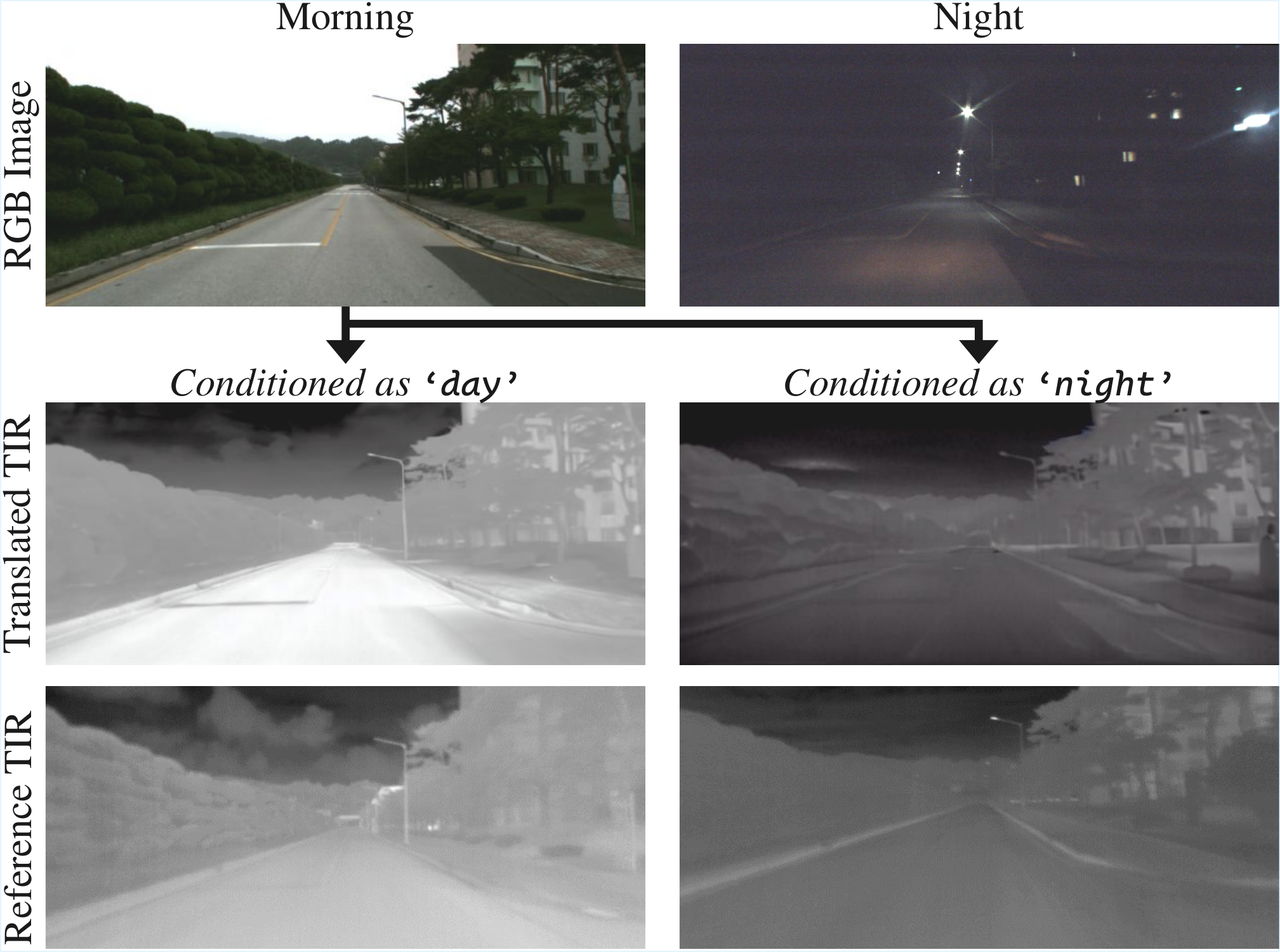}
 \caption{\textbf{Day-Night Controllability (Thermal Inversion)} Input RGB and ground truth TIR images are taken from STheReO \cite{sthereo}. TherA simulates the diurnal cycle from a single RGB image, matching the ground truth thermal characteristics.}
  \label{fig:day_night1}
   \vspace{-5mm}
\end{figure}

A further contribution of TherA is providing the first framework for controllable RGB-to-\ac{TIR} translation. \cref{controllability_results} shows both text- and reference-guided control, realistically rendering various weathers (\cref{controllability_results2}) and time-of-day (\cref{fig:day_night1}) with accurate changes to global contrast and reflections, as well as object-level control of vehicles, toggling specific cars between being active and parked (\cref{controllability_results3}) with heat correctly depicted and emitted from wheels and exhausts.


Additionally, TherA provides a functional advantage in addressing a key limitation of \ac{TIR} imaging, thermal inversion \cite{shin2019sparse, lee2024caltech-cart}, where diurnal temperature shifts cause large day-night discrepancies in apparent \ac{TIR} intensity. Translated TIR images must account for this discrepancy, but generating realistic night-time pseudo-\ac{TIR} image is particularly difficult as night-time RGB inputs are dominated by noise and low visibility, offering little reliable signal from which to infer thermal structure.

TherA’s controllability directly mitigates this limitation. As shown in \cref{fig:day_night1}, it leverages a clean daytime RGB input and, with a simple text prompt, synthesizes a physically plausible nighttime \ac{TIR} image. This bypasses the poor nighttime RGB dependency of prior methods and enables scalable creation of all-weather, all-time pseudo-\ac{TIR} datasets, effectively alleviating the data scarcity that has long constrained thermal imaging.

\subsection{Ablation Studies}

\subsubsection{Importance of TherA-VLM Thermal Embedding}
We conduct an ablation study (\cref{translation_results_ablation}) to validate our core design choice: conditioning on the thermal embedding of our TherA-VLM.
We test four progressive variants:
(\emph{i}) Baseline: InstructPix2Pix \cite{brooks2023instructpix2pix} trained on R2T2.
(\emph{ii}) LLaVA: A generic VLM \cite{liu2024improvedllava}.
(\emph{iii}) LLaMA: Our TherA-VLM's \emph{text output} (the schema) as a condition.
(\emph{iv}) TherA: Our TherA-VLM's \emph{thermal embedding} as a condition.

Simply scaling data or using a generic VLM yields poor results.
Using the canonicalized text provides a marginal improvement, but the model remains RGB-centric.
Only TherA, which conditions on the rich, latent hidden states, achieves a massive leap in performance (e.g., +4.66 dB PSNR on M3FD over LLaMA). This validates that our thermal-aware \emph{vision-language reasoning}, not just the text, is key to physical realism.

\begin{table}[h]
\centering
\vspace{-2mm}
\caption{Ablation results. 
Best results are highlighted in \textbf{bold} and second best are \underline{underlined}.}
\vspace{-2mm}
\resizebox{\columnwidth}{!}{%
\begin{tabular}{l|cccc|cccc}
\toprule
\multicolumn{1}{c|}{\multirow{2}{*}{\textbf{Method}}} &
\multicolumn{4}{c|}{\textbf{M3FD} \cite{m3fd}} &
\multicolumn{4}{c}{\textbf{FLIR} \cite{fliradas}} \\
\cmidrule(lr){2-5} \cmidrule(lr){6-9}
 & 
PSNR$\uparrow$ & SSIM$\uparrow$ & FID$\downarrow$ & LPIPS$\downarrow$ &
PSNR$\uparrow$ & SSIM$\uparrow$ & FID$\downarrow$ & LPIPS$\downarrow$ \\
\midrule
InstructPix2Pix \cite{brooks2023instructpix2pix} & 12.40 & 0.32 & 203.01 & 0.45 & 13.38 & 0.35 & 133.51 & 0.46 \\
LLaVA & 11.85 & 0.47 & \underline{135.12} & \underline{0.37} & 14.25 & 0.40 & \textbf{101.82} & \underline{0.39} \\
Lavi-LLama & \underline{13.23} & \underline{0.51} & 142.37 & 0.41 & \underline{14.65} & \underline{0.42} & 113.75 & 0.40 \\
\textbf{TherA} & \textbf{18.24} & \textbf{0.65} & \textbf{105.52} & \textbf{0.29} & \textbf{16.56} & \textbf{0.49} & \underline{112.93} & \textbf{0.38} \\ 
\bottomrule
\end{tabular}%
}
\label{translation_results_ablation}
\vspace{-5mm}
\end{table}

\subsubsection{CFG Scales}
We analyze the impact of the dual \ac{CFG} scales for image guidance ($s_V$) and VLM guidance ($s_S$) in \Cref{sec:ablation_cfg}. We confirmed that optimal translation comes from minimizing RGB-latent reliance ($s_V=0.5$) and leveraging TherA-VLM’s stronger thermal prior ($s_S=1.5$).

\subsubsection{Downstream Task Evaluation}

To assess the practical value of our generated thermal data, we evaluate two downstream tasks: \textbf{thermal segmentation} and \textbf{RGB-TIR image matching}. As shown in \Cref{sec:ablation_segmentation} and \Cref{sec:ablation_matching}, pretraining thermal segmentation and RGB-TIR image matching models using TherA's pseudo-TIR images enhance the performance of both tasks, in comparison to other image translation baselines.




\subsubsection{Comparison with Reference-guided Models}

We compared TherA's reference-guided image translation performance with existing example-guided image translation models. As illustrated in \Cref{sec:ablation_referenceguided}, TherA is the only model that depicts thermally realistic characteristics. 
\section{Conclusion}
\label{sec:conclusion}

We present TherA, a controllable RGB-to-\ac{TIR} translation framework conditioned by a thermal-aware vision-language model (TherA-VLM).
TherA generates thermally plausible images that capture scene-level physics while allowing explicit control over environmental states and object activity.
Comprehensive evaluations show that TherA consistently improves perceptual and structural fidelity over prior methods, producing radiometrically coherent and controllable pseudo-\ac{TIR} images. 

\noindent \textbf{Limitations}: Currently, our method operates on relative thermal imagery, where pixel intensities represent normalized or contrast-based temperature differences rather than absolute radiometric values. However, we believe this framework opens new directions for scalable pseudo-thermal data generation, multispectral perception, and thermally grounded vision-language modeling.

{
    \small
    \bibliographystyle{IEEEtranN}
    \bibliography{main}
}
\clearpage
\setcounter{page}{1}
\renewcommand{\thesection}{\Alph{section}}
\renewcommand{\thesubsection}{\thesection.\arabic{subsection}}
\setcounter{section}{0}

\maketitlesupplementary

\section{Additional Implementation Details}
\label{sec:additional_implementation_details}

\subsection{Datasets}
\begin{figure}
    \centering
    \begin{subfigure}[t]{\columnwidth}
        \centering
        \includegraphics[width=0.9\columnwidth]{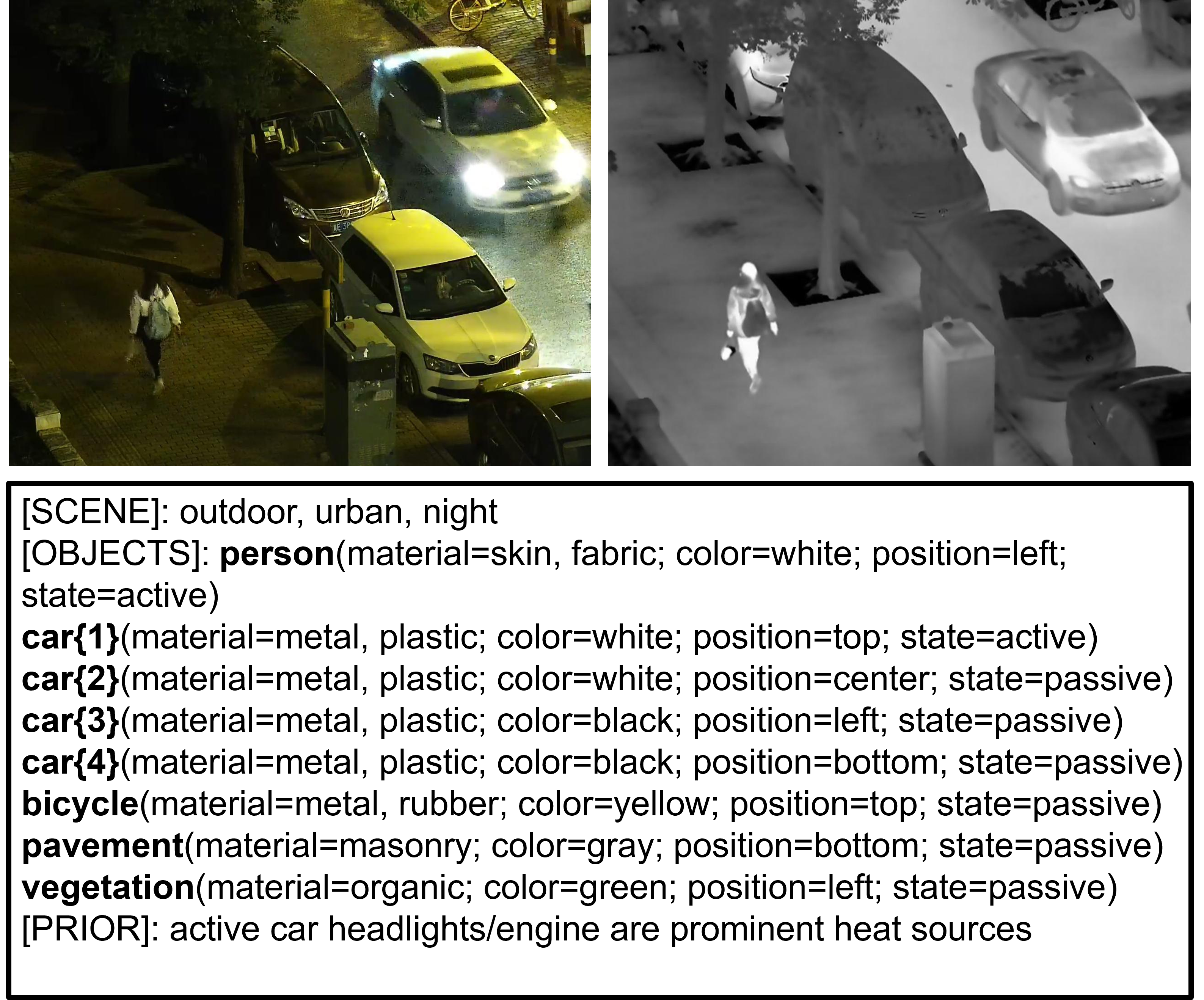} 
        \caption*{(a) RGB-TIR-text example from LLVIP \cite{llvip}}
    \end{subfigure}
    \hfill
    \begin{subfigure}[t]{\columnwidth}
        \centering
        \includegraphics[width=0.9\columnwidth]{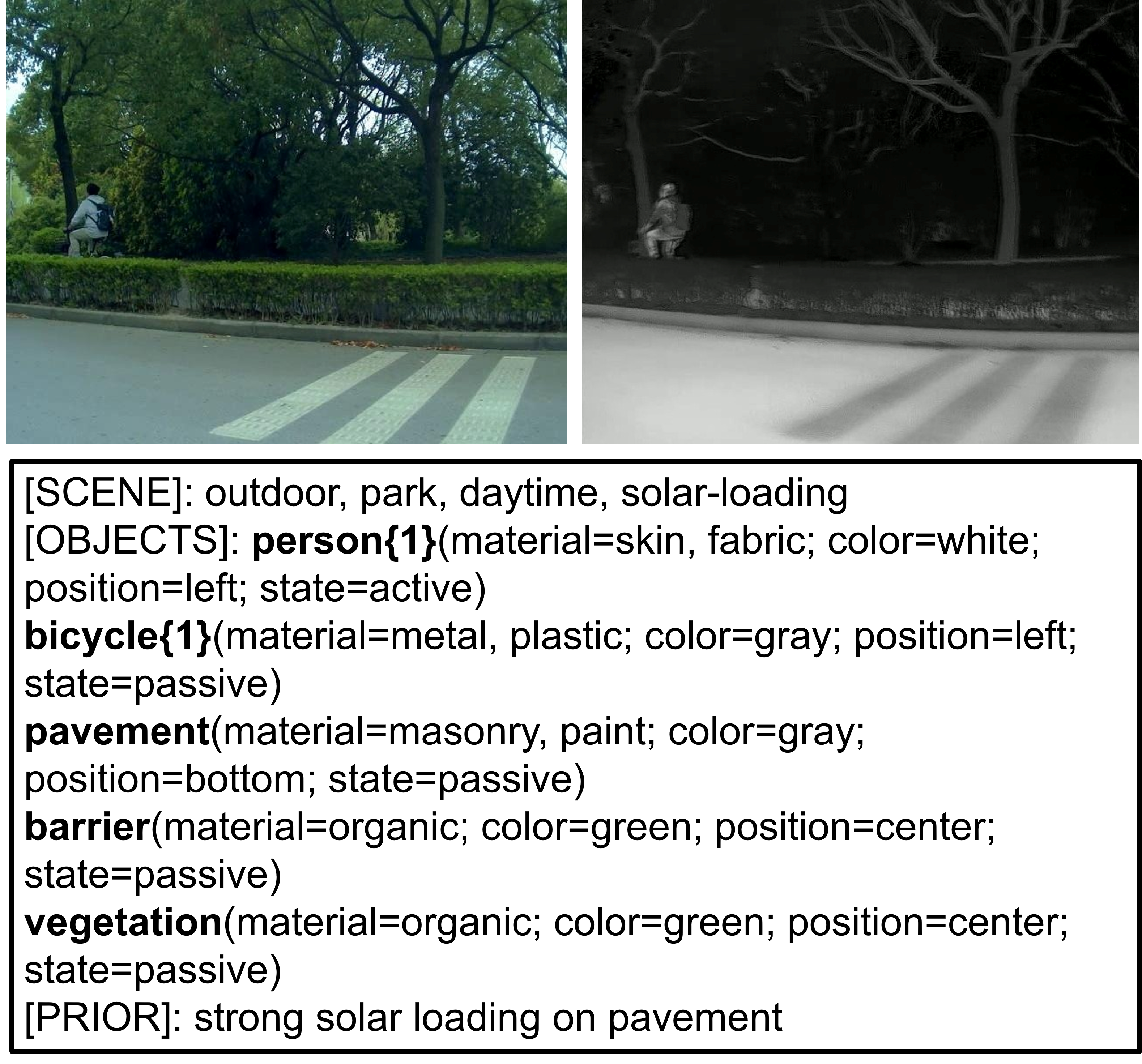} 
        \caption*{(b) RGB-TIR-text example from SMOD \cite{sjtu}}
    \end{subfigure}
        \hfill
    \begin{subfigure}[t]{\columnwidth}
        \centering
        \includegraphics[width=0.9\columnwidth]{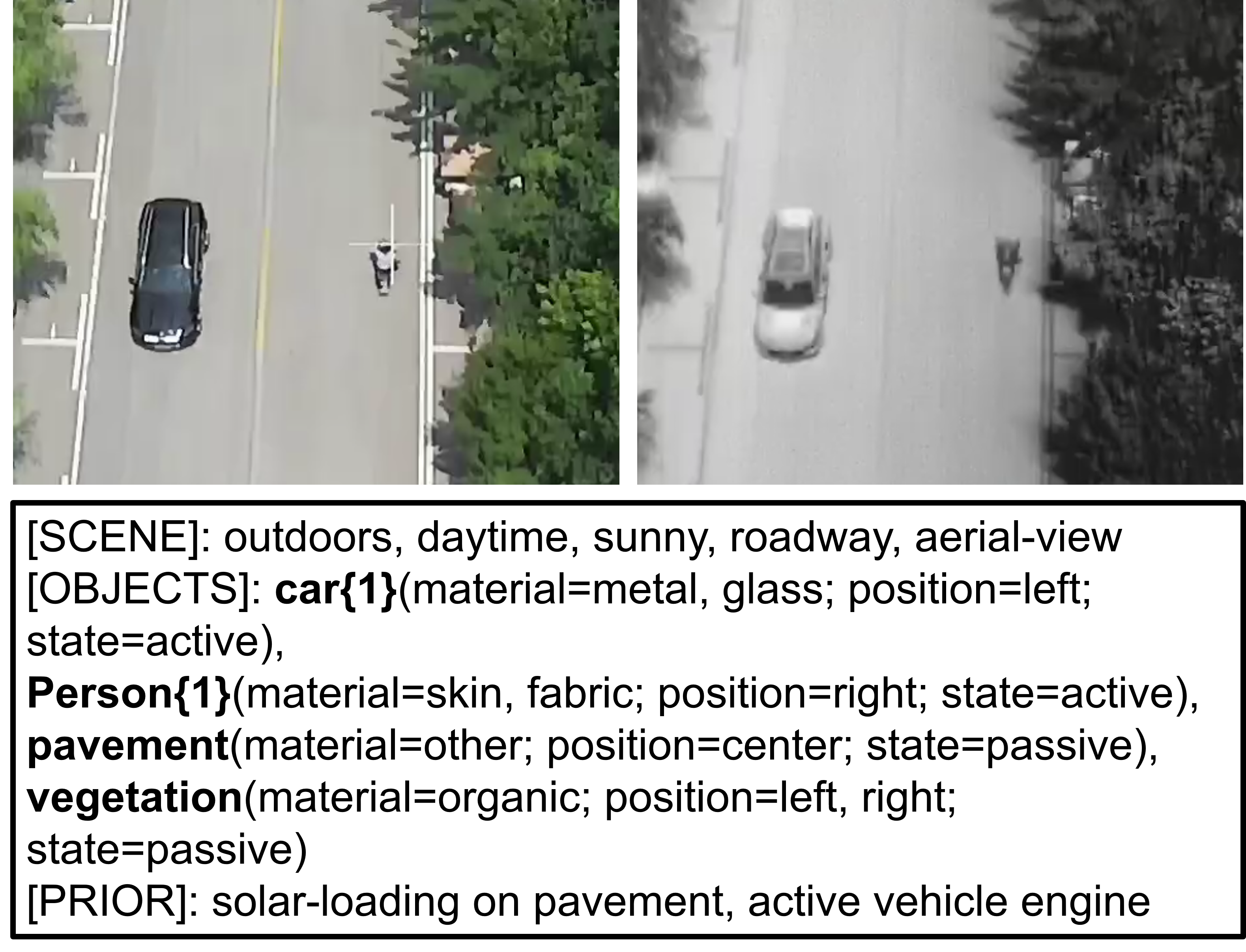} 
        \caption*{(b) RGB-TIR-text example from AVIID \cite{aviid}}
    \end{subfigure}

     \caption{Examples of RGB--TIR pairs in R2T2 dataset.}
    \label{fig:canonical-schema-examples}
\end{figure}

\paragraph{R2T2.}
R2T2 is a RGB--TIR--text dataset that we designed  specifically to train TherA-VLM and our diffusion-based translation model. R2T2 provides spatially aligned RGB--TIR image pairs together with canonicalized thermal-aware captions that are used to train TherA-VLM and the VLM-conditioned diffusion model (see \cref{sec:R2T2_curation} for curation details). In total, R2T2 contains 112{,}970 RGB--TIR--text triplets. We use all available RGB--TIR pairs to train our models, with the exception of samples that overlap with the VTMOT-test, FLIR, and M3FD sets. R2T2 serves as the primary supervision source for both TherA-VLM and our diffusion-based translation model, and all models used in our zero-shot experiments are trained exclusively on R2T2. Example data samples are provided in  \cref{fig:canonical-schema-examples}

\begin{table}[!h]
\centering
\caption{Constituent datasets of R2T2 and their statistics. ``Sample Number (Translation)'' denotes the number of RGB--TIR pairs used for training the translation model. The total across all sources is 112{,}970 pairs.}
\label{tab:R2T2-stats}
\resizebox{\columnwidth}{!}{%
\begin{tabular}{c|cccc}
\hline \hline
\textbf{Dataset} & \textbf{\begin{tabular}[c]{@{}c@{}}Sample Number\\ (Translation)\end{tabular}} & \textbf{\begin{tabular}[c]{@{}c@{}}Average\\ Resolution\end{tabular}} & \textbf{Viewpoint} & \textbf{Split} \\ \hline \hline
AVIID \cite{aviid}            & 683                                                                            & 480×480                                                               & Aerial             & Train          \\
CAMEL \cite{camel}            & 3481                                                                           & Diverse                                                               & CCTV               & Train          \\
FLIR \cite{fliradas}             & 5141                                                                           & 640×512                                                               & Driving            & Finetune/Test  \\
KAIST \cite{kaist-mdpd}            & 6457                                                                           & 640×512                                                               & Driving            & Train          \\
LLVIP \cite{llvip}            & 15488                                                                          & 640×512                                                               & CCTV               & Train          \\
M3FD \cite{m3fd}             & 4200                                                                           & 512×384                                                               & Driving            & Finetune/Test  \\
METU-VisTIR \cite{xoftr-metu-vistir}      & 408                                                                            & 640×512                                                               & Ego-View           & Train          \\
MSRS \cite{msrs}             & 1340                                                                           & 640×480                                                               & Driving            & Train          \\
NSAVP \cite{nsavp}            & 23520                                                                          & 640×512                                                               & Driving            & Train          \\
SMOD \cite{sjtu}             & 4553                                                                           & 640×512                                                               & Driving            & Train          \\
VIVID \cite{vivid++}            & 23012                                                                          & 640×512                                                               & Driving            & Train          \\
VTMOT \cite{vtmot}            & 25079                                                                          & Diverse                                                               & Ego-view           & Train/Val      \\ \hline \hline
\end{tabular}%
}
\end{table}

Table~\ref{tab:R2T2-stats} summarizes the constituent datasets used to build R2T2, including the number of RGB--TIR pairs contributing to translation training, their typical resolutions, viewpoints (driving, CCTV, aerial, ego-view), and data split. This composition provides a diverse mixture of viewpoints and environments, while ensuring that samples from FLIR, M3FD, and the VTMOT-test split used for benchmark finetuning and evaluation are clearly separated from those used for training.

\paragraph{Benchmark datasets and evaluation protocol.}
For benchmark finetuning and evaluation, we use the M3FD~\cite{m3fd}, FLIR~\cite{fliradas}, and CART~\cite{lee2024caltech-cart} datasets. Following the non-overlapping splits provided in prior work~\cite{ran2025diffv2ir,flir-split}, we use 3{,}550 training images and 650 test images for M3FD, and 4{,}128 training images and 1{,}013 test images for FLIR. For CART, we adopt the official test split of 280 images from~\cite{lee2024caltech-cart} and do not use any CART images for training.

For M3FD and FLIR, we consider two evaluation regimes: (i) \emph{zero-shot}, where the model is trained only on R2T2 and evaluated directly on the corresponding test split, and (ii) \emph{finetuned}, where the R2T2-pretrained model is further finetuned on the respective training split and evaluated on the held-out test split. For CART, we report only zero-shot results: the model is trained on R2T2 and evaluated on the CART test split without any additional finetuning.

\subsection{TherA-VLM}

We fine-tune LLaVA v1.5 (7B)~\cite{liu2024improvedllava} with a CLIP ViT-L/14-336 vision encoder~\cite{clip} for thermally aware instruction tuning. The model is trained on the R2T2 dataset with a 70:30 train/validation split. We use the Vicuna v1 conversation template, and compute a standard next-token cross-entropy loss over assistant tokens while masking out image tokens and user tokens from the loss.

For fine-tuning, we attach LoRA adapters (rank 128, $\alpha = 256$) to the linear layers of the language model. The LLM backbone and vision encoder remain frozen; only the LoRA parameters and the image-to-text projector weights are updated. We optimize with AdamW (weight decay $0.0$), using a cosine learning-rate schedule with a warmup ratio of $0.03$ and a maximum sequence length of 2048 tokens. The base learning rate for the language model is set to $2 \times 10^{-4}$ and the projector learning rate to $2 \times 10^{-5}$. Training is run for one epoch with a per-GPU batch size of 16 and gradient accumulation of 2 across 4 GPUs, resulting in an effective batch size of 128.

\subsection{Image Translation}

For image translation, input RGB and TIR images are resized to $256 \times 256$, converted to tensors, and normalized to $[-1, 1]$. We use the Stable Diffusion v1.4 VAE (frozen) to encode both RGB and TIR frames into latent representations. The UNet operates on an 8-channel input formed by concatenating noisy TIR latents with RGB latents along the channel dimension. We initialize the UNet weights from Stable Diffusion v1.4 and the text-to-image adapter (TE-Adaptor) from the LLaMA–UNet adapter of~\cite{zhao2024bridging}. The VAE and the VLM model remain frozen; we update the UNet and TE-Adaptor parameters.

We train the diffusion model with AdamW on the UNet and adapter parameters, using decoupled weight decay ($10^{-4}$ on weight tensors), betas $(0.9, 0.999)$, and $\epsilon = 10^{-8}$. We apply a cosine learning-rate schedule with warm-up and set the peak learning rate to $1 \times 10^{-4}$. Training is performed for 100 epochs with a per-GPU batch size of 50 and gradient accumulation of 1 on 4 NVIDIA A6000 GPUs. We use \texttt{bfloat16} mixed precision with \texttt{TF32} enabled and apply gradient clipping with a maximum global norm of $0.5$. The training objective is the standard diffusion noise-prediction MSE loss. During training, we employ classifier-free guidance (CFG) dropout by independently dropping the unconditional, text, and image conditions with probability 0.1 to improve robustness.

\section{R2T2 Curation}
\label{sec:R2T2_curation}

\subsection{Thermal-aware Text Generation}

\subsubsection{From Radiometric Chain to Semantic Schema}

Although we only observe thermal images that display relative temperature in our datasets (8-bit, normalized TIR frames that preserve the ordering of temperatures, but not their absolute values), the formation of these images is still governed by the standard radiometric chain \cite{vollmer2018infrared}. The radiometric chain states that the irradiance measured by an TIR detector can be written as
\begin{equation}
\Phi_{\text{det}}
=
\tau_{\text{atm}}\,\varepsilon\,\Phi^{bb}(T_{\text{obj}})
+
\tau_{\text{atm}}(1-\varepsilon)\,\Phi_{\text{amb}}
+
(1-\tau_{\text{atm}})\,\Phi_{\text{atm}},
\end{equation}
where $\varepsilon$ is the material emissivity, $T_{\text{obj}}$ is the object temperature, $\Phi^{bb}(\cdot)$ denotes black-body emission, and $T_{\text{amb}}, T_{\text{atm}}, \tau_{\text{atm}}$ control ambient and atmospheric contributions.

Because our TIR images are normalized, we cannot directly recover absolute values of $\varepsilon$, $T_{\text{obj}}$, or $T_{\text{amb}}$ from pixel intensities. However, the \emph{relative} contrast and structure observed in TIR still arise from variations in these physical factors. Our canonical schema is designed as a high-level, discretized parameterization of this radiometric chain:
\begin{itemize}
    \item \textbf{Material} groups objects into emissivity classes, providing a coarse proxy for $\varepsilon$.
    \item \textbf{Heat Emission States} encodes the relative thermal state of an object (e.g., parked vs.\ driving car), acting as a proxy for $T_{\text{obj}}$.
    \item \textbf{Scene} (time of day, weather, environment) captures conditions that influence $T_{\text{amb}}, T_{\text{atm}}$ and $\tau_{\text{atm}}$ (e.g., solar loading on roads, nighttime cooling).
    \item \textbf{Object} category supplies priors over typical spatial heat distributions (e.g., hot engine region vs.\ cooler body, warm human torso vs.\ cooler background).
\end{itemize}

Without these four components, key terms in the radiometric chain become effectively unobserved from RGB alone, making the RGB-to-TIR mapping severely underdetermined. By explicitly modeling these components, we obtain a physically motivated, semantically interpretable discretization of the radiometric chain that provides exactly the information needed for stable conditioning of the diffusion model, while remaining compatible with the relative (normalized) TIR images used in practice.

\subsubsection{Canonical Schema Generation}

Generating text only from an RGB image does not reveal how the scene appears in the thermal infrared domain. Since our goal is to train a \emph{thermal-aware} vision–language model, we construct supervision using a multimodal reasoning model, Gemini~2.5~Pro~\cite{team2024gemini}, tied explicitly to our radiometric schema (Sec.~\ref{sec:R2T2_curation}). Given paired RGB–TIR inputs, we query Gemini with a structured, slot-based prompt that returns three fields: \texttt{SCENE}, \texttt{OBJECTS}, and \texttt{PRIOR}. The prompt constrains object names, materials, colors, and states to a fixed vocabulary, enforces a specific format (e.g., \texttt{name(material=...; color=...; position=...; state=active|passive)}), and asks Gemini to report only high-confidence thermal attributes. The \texttt{SCENE} and \texttt{OBJECTS} fields describe time of day, weather, environment, object categories, materials, and heat emission (active/passive) states, whereas \texttt{PRIOR} is reserved for thermal context that is typically not recoverable from RGB alone (e.g., solar loading on pavement, whether parked cars are active or passive). The full prompt is available in \cref{fig:gemini-prompt}

\paragraph{Handling non-observable thermal states: PRIOR.}

Even with paired RGB–TIR inputs, some thermal factors are ambiguous from RGB appearance (e.g., accumulated solar loading, or whether a vehicle is currently running). If these factors are omitted from the conditioning text while they are clearly present in the TIR image, the model is trained under a semantic–image mismatch, which encourages either hallucinated thermal patterns or ignoring the text altogether. To mitigate this, we introduce the \texttt{PRIOR} slot, which is instantiated from TIR evidence during data curation (e.g., “sun-heated pavement”, “parked cars passive”, “electronics off”). During training and inference, we append the PRIOR phrases to the user instruction that conditions TherA-VLM, leveraging LLaVA’s conversational interface. In practice, PRIOR serves as a physically motivated control knob: it aligns the conditioning text with the ground-truth TIR during training and later enables explicit text-guided control of thermal behaviour in the translation model.


\begin{figure}[h]
    \centering
    \begin{subfigure}[t]{\columnwidth}
        \centering
        \includegraphics[width=\columnwidth]{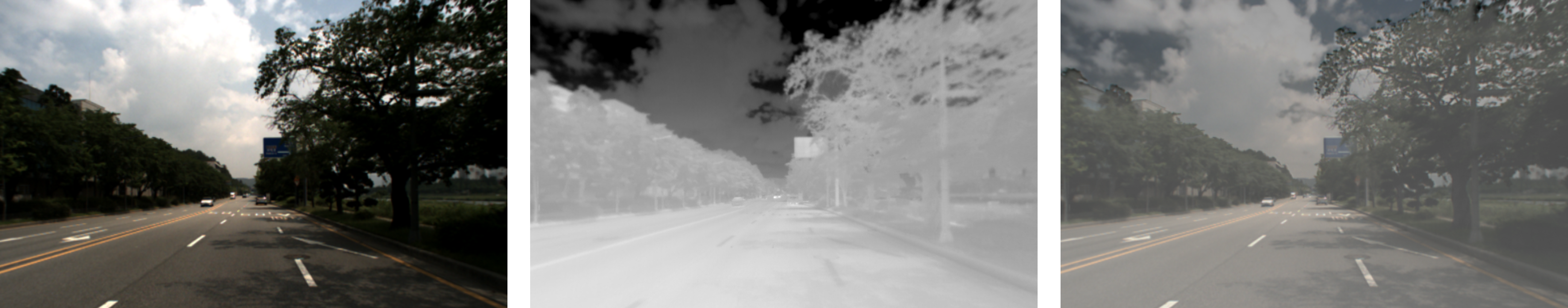} 
        \caption*{(a) Valid alignment example}
    \end{subfigure}
    \hfill
    \begin{subfigure}[t]{\columnwidth}
        \centering
        \includegraphics[width=\columnwidth]{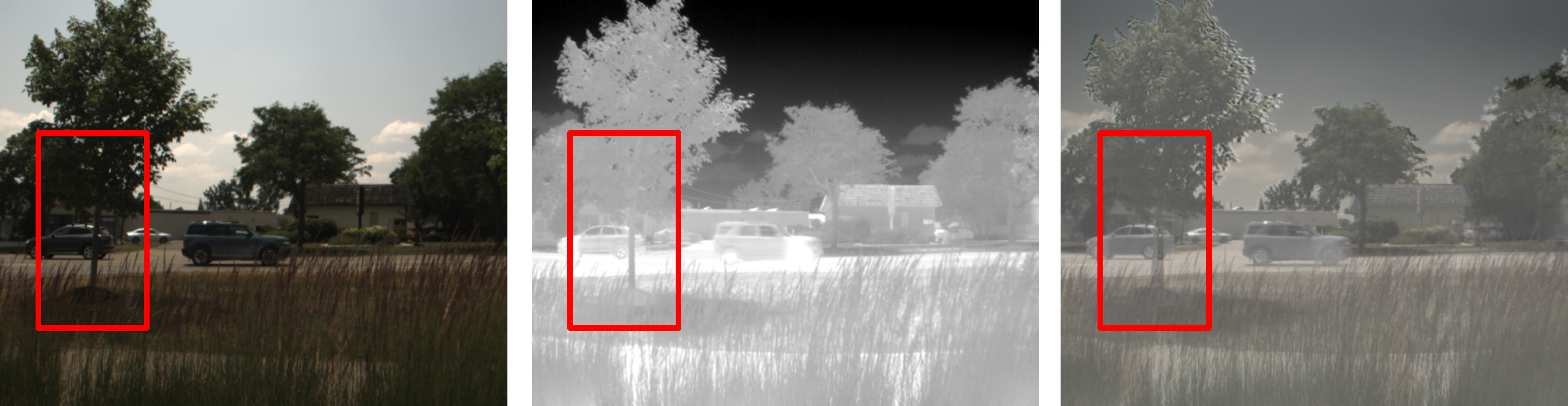} 
        \caption*{(b) Invalid alignment example}
    \end{subfigure}

    \caption{
        \textbf{Examples of pseudo-aligned RGB--TIR pairs.} We use alpha-blended overlays (right) of RGB (left) and TIR (middle) to visually screen for residual misalignment, and keep only pairs with minimal visible parallax.
    }
    \label{fig:pseudo-align-examples}
    \vspace{-4mm}
\end{figure}

\paragraph{Canonicalizing free-form reasoning.}
Raw free-form descriptions from Gemini contain a large, redundant, and noisy vocabulary, which led to unstable conditioning in early experiments. We therefore canonicalize Gemini’s reasoning into a compact, physics-informed schema. Scene and object labels are mapped to the COCO \emph{things/stuff} taxonomy~\cite{caesar2018coco}, and material groups are mapped to emissivity classes defined in HADAR~\cite{bao2023hadar}. Each object instance is annotated with a canonical material group, color, position (left/right/top/bottom), and heat emission state (\emph{active} or \emph{passive}), yielding a vocabulary of 23 object classes, 13 material groups, 14 colors, 4 positions, and 2 heat emission states. We first instruct Gemini to follow this schema and then apply a post-hoc LLM pass that merges synonyms into canonical tokens (e.g., “sedan”/“SUV” $\rightarrow$ \texttt{car}; “asphalt”/“road surface” $\rightarrow$ \texttt{road}). Finally, because we operate on RGB–TIR pairs, we run a consistency check in which the LLM flags and corrects contradictions between the canonicalized text and the observed TIR patterns (e.g., an object described as “cool” but appearing hotter than its surroundings). Although PRIOR and active/passive states may still be occasionally mispredicted, the restricted vocabulary and consistency filtering substantially reduce noise and yield stable, thermally grounded captions, which we find to be markedly more effective than raw free-form descriptions in our ablations.

\begin{figure*}[t]
\centering
\begin{minipage}{0.95\textwidth}
\footnotesize\ttfamily
TASK: Describe how the RGB scene would appear in TIR. Use only high-confidence facts; abstain otherwise.

INPUTS: RGB\_IMAGE, THERMAL\_IMAGE

OUTPUT: Single text block. No extra text outside fields.

SCENE: Identify the general environment of the weather that could influence the thermal
characteristics of the image. These could include locations, weather, time of the day,
indoor/outdoors, season. Refer to heat intensities of the thermal image and the appearance
of RGB images to make reasonable deductions. Only output deductions that you are confident
with. State them by keywords separated by commas. Omit uncertain items.

OBJECTS: ITEM FORMAT: name(material=...; color=...; position=left|right|top|bottom|center;
state=active|passive).

THINGS (instanceable):
  \{person, people, car, truck, motorcycle, bicycle, other\_vehicle,
   building\_equipment, streetlight, sensor,
   furniture, electronics, object, animal, potted\_plant\}

STUFF (non-instanceable; record ONCE only):
  \{building, pavement, barrier, structure, street-furniture, vegetation, traffic-sign\}

MATERIALS (allowed):
  \{metal, glass, plastic, rubber, masonry, organic, fabric, skin, air, paint, light, water, other\}

COLORS (allowed):
  \{white, black, gray, brown, cream, red, blue, yellow, green, orange, purple, pink, navy, na\}

STATE RULES: active = self-heating (engines/electronics/living bodies);
             passive = environment/solar only.
Vehicles: active if hotspot in hood/exhaust/wheels; side-parked + no hotspot -> passive.

GROUPING (THINGS only): split\_by\_state -> if (center\_dist < 0.05*diag OR IoU > 0.10 OR occlusion)
                        -> plural (cars, people) else index as name\{1\}, name\{2\}, ...

STUFF: no indices unless that part is a Prominent heat source
-- OMIT: small \& low-contrast \& thermally irrelevant
-- UNCERTAIN but relevant: map to nearest allowed name; material=other if unsure
-- EXCLUDE: artifacts/flares/blurs from OBJECTS

PRIOR:
- Short phrases of thermal context NOT visible from RGB (e.g., solar-loading,
  sky<<ground, parked cars passive, electronics off).
- Mention active/passive classes if relevant.

Notes:
- Use ONLY allowed tokens for names/materials/colors
- Prefer omission or "other" over speculation
- Output exactly in the field order above
\end{minipage}
\caption{\textbf{Gemini Prompt.} Prompt used to query Gemini~2.5~Pro for thermal-aware captions over our schema fields (SCENE, OBJECTS, PRIOR).}
\label{fig:gemini-prompt}
\end{figure*}


\newpage


    

\subsection{Pseudo-aligned RGB--TIR Pairs}

Many public RGB--TIR datasets are time-synchronized but not spatially aligned, which makes them unsuitable for direct pixel-level supervision. To increase the diversity of our training data, we convert such datasets into \emph{pseudo-aligned} RGB--TIR pairs. Our goal is not to obtain perfect pixel-wise registration, but to identify frames where the residual misalignment is small enough to be useful for training. These pseudo-aligned pairs are used only for training the translation model; all quantitative evaluations rely exclusively on natively aligned benchmarks (M3FD, FLIR, CART).

For each synchronized RGB--TIR frame, we estimate cross-modal correspondences with MINIMA~\cite{minima}, and fit a global warp that maps the RGB image into the TIR view. For datasets where camera extrinsics and LiDAR scans are available, we first apply the provided calibration to coarsely rectify the views and then run the same correspondence-based refinement. Because the scenes are not strictly planar and the sensors have a non-zero baseline, the resulting pairs are only approximately registered; we therefore explicitly treat them as pseudo-aligned.

To filter out pairs with unacceptable parallax or warping artifacts, we adopt a simple human-in-the-loop check. For each candidate pair, we generate an alpha-blended RGB--TIR overlay (0.5/0.5) and ask three annotators to label it as \emph{accept} or \emph{reject}, using a small set of reference examples for guidance (see \cref{fig:pseudo-align-examples}). Only pairs with majority \emph{accept} votes are retained. Applying this procedure across several RGB--TIR datasets~\cite{vivid++,nsavp,camel} yields 50{,}013 additional pseudo-aligned pairs used solely as extra training data.

\newpage
\section{Additional Qualitative Results}
\label{sec:qualitative}

\subsection{TherA-VLM Examples}
\label{sec:thera-vlm_example}

\begin{figure*}
    \centering
    \begin{subfigure}{\textwidth}
        \centering
        \includegraphics[width=0.78\textwidth]{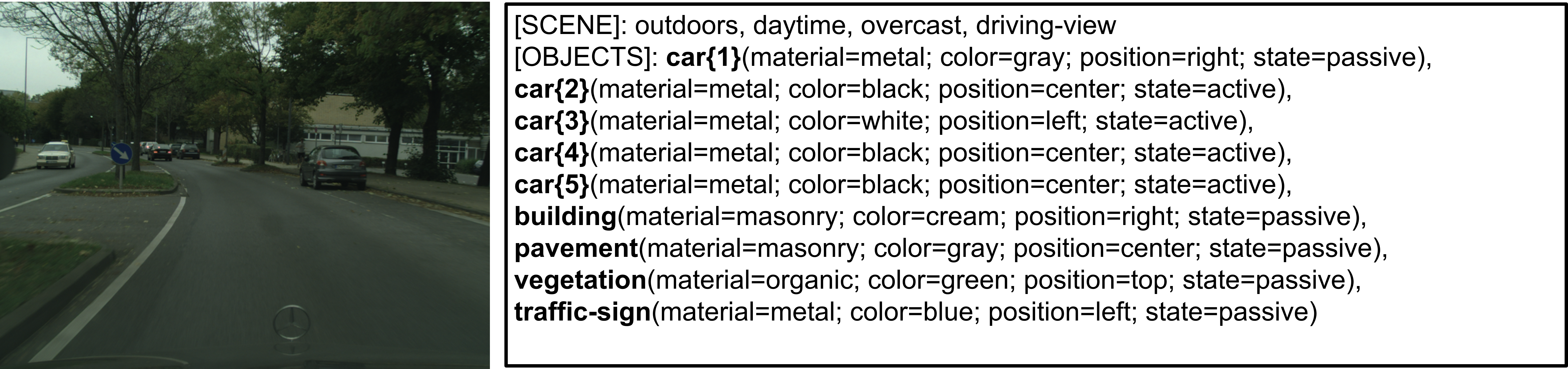}
        \caption{Example TherA-VLM Output: Cityscapes \cite{cordts2016cityscapes}}
        \label{fig:thera-vlm-cityscapes}
    \end{subfigure}
    \vspace{1.5mm} 
    \begin{subfigure}{\textwidth}
        \centering
        \includegraphics[width=0.78\textwidth]{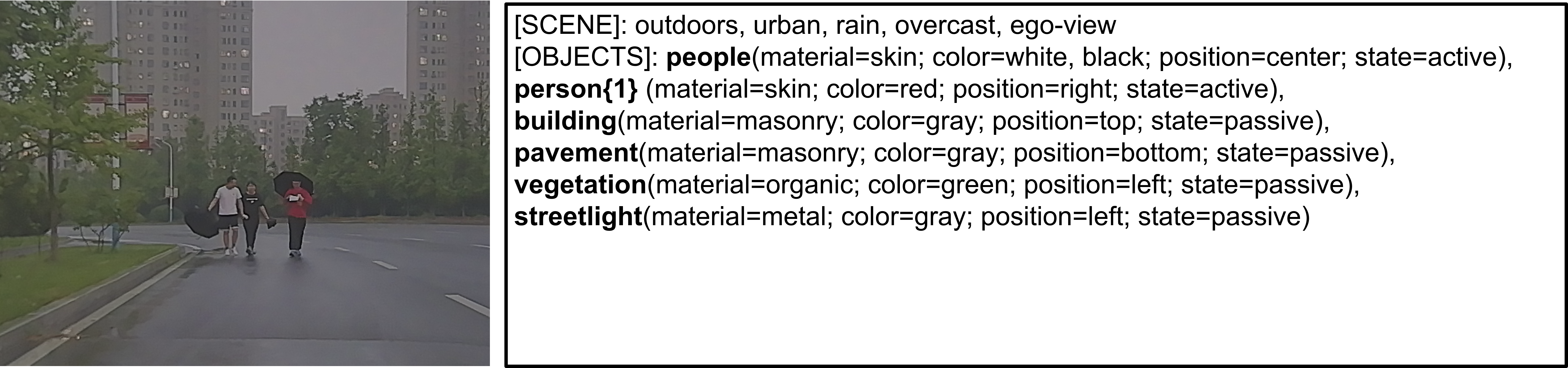}
        \caption{Example TherA-VLM Output: FMB \cite{fmb}}
        \label{fig:thera-vlm-fmb}
    \end{subfigure}
    \begin{subfigure}{\textwidth}
        \centering
        \includegraphics[width=0.78\textwidth]{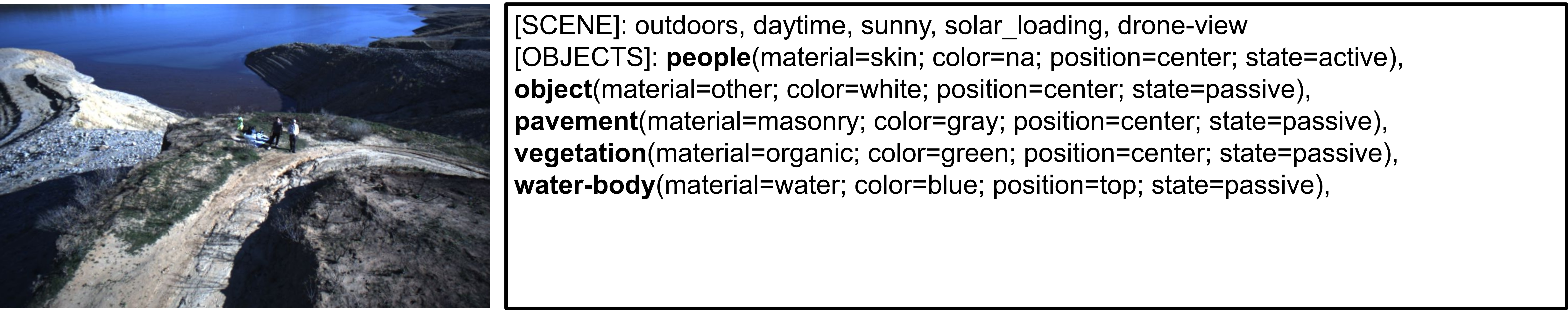}
        \caption{Example TherA-VLM Output: CART \cite{lee2024caltech-cart}}
        \label{fig:thera-vlm-cart}
    \end{subfigure}
    \vspace{1.5mm} 
    \begin{subfigure}{\textwidth}
        \centering
        \includegraphics[width=0.78\textwidth]{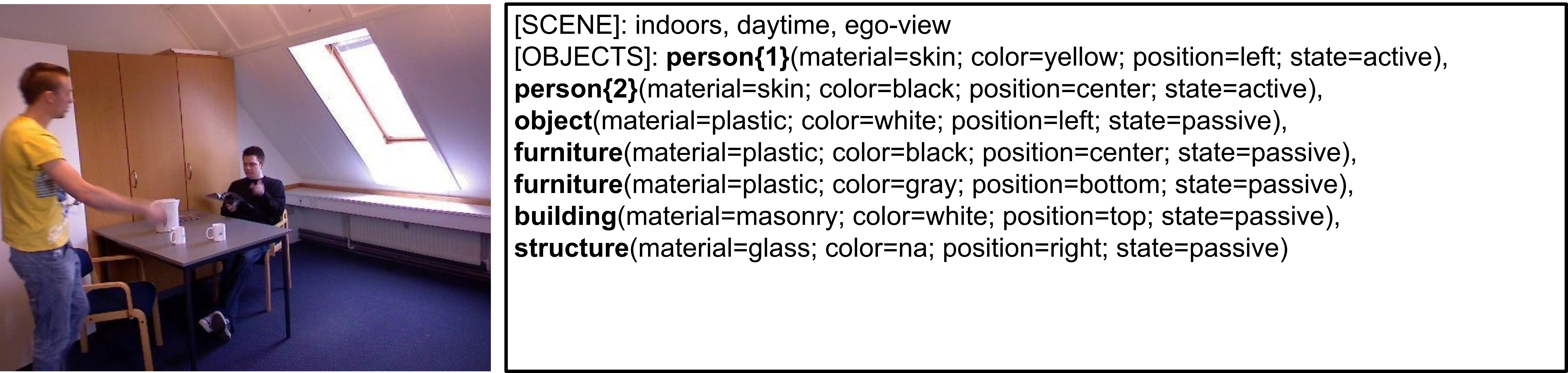}
        \caption{Example TherA-VLM Output: Trimodal \cite{trimodal}}
        \label{fig:thera-vlm-trimodal}
    \end{subfigure}
    \begin{subfigure}{\textwidth}
        \centering
        \includegraphics[width=0.78\textwidth]{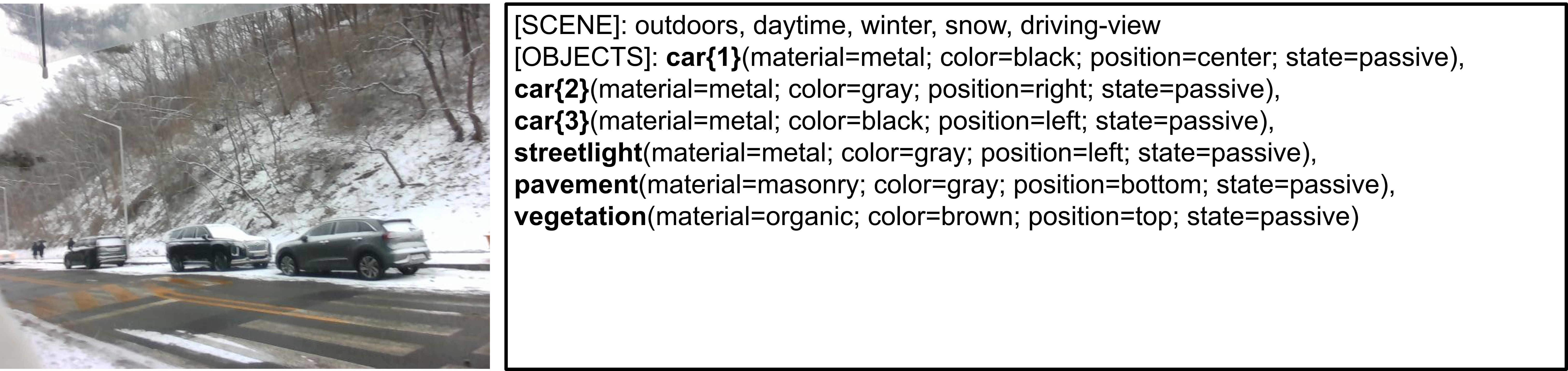}
        \caption{Example TherA-VLM Output: Custom}
        \label{fig:thera-vlm-snow}
    \end{subfigure}
    \vspace{-4mm}
    
    \caption{
        \textbf{TherA-VLM Examples.} Example output results for TherA-VLM across different benchmarks
    }
    \label{fig:TherA-VLM-outputs}
    \vspace{-4mm}
\end{figure*}

Figure~\ref{fig:TherA-VLM-outputs} illustrates the generated language output of TherA-VLM on unseen images. Here, images from diverse scenes, including urban driving, aerial terrain, rainy streets, snowy roads, and indoor environments are presented.   
Across all cases, the model produces coherent SCENE descriptions (e.g., rain, snow, solar loading, indoor vs.\ outdoor) and assigns materials and active/passive states to objects (vehicles, people, vegetation, buildings, furniture) in a manner consistent with typical thermal behavior, providing physically grounded conditioning signals for the translation model.

\newpage

\begin{figure*}[t]
 \centering
 \includegraphics[width=0.83\textwidth]{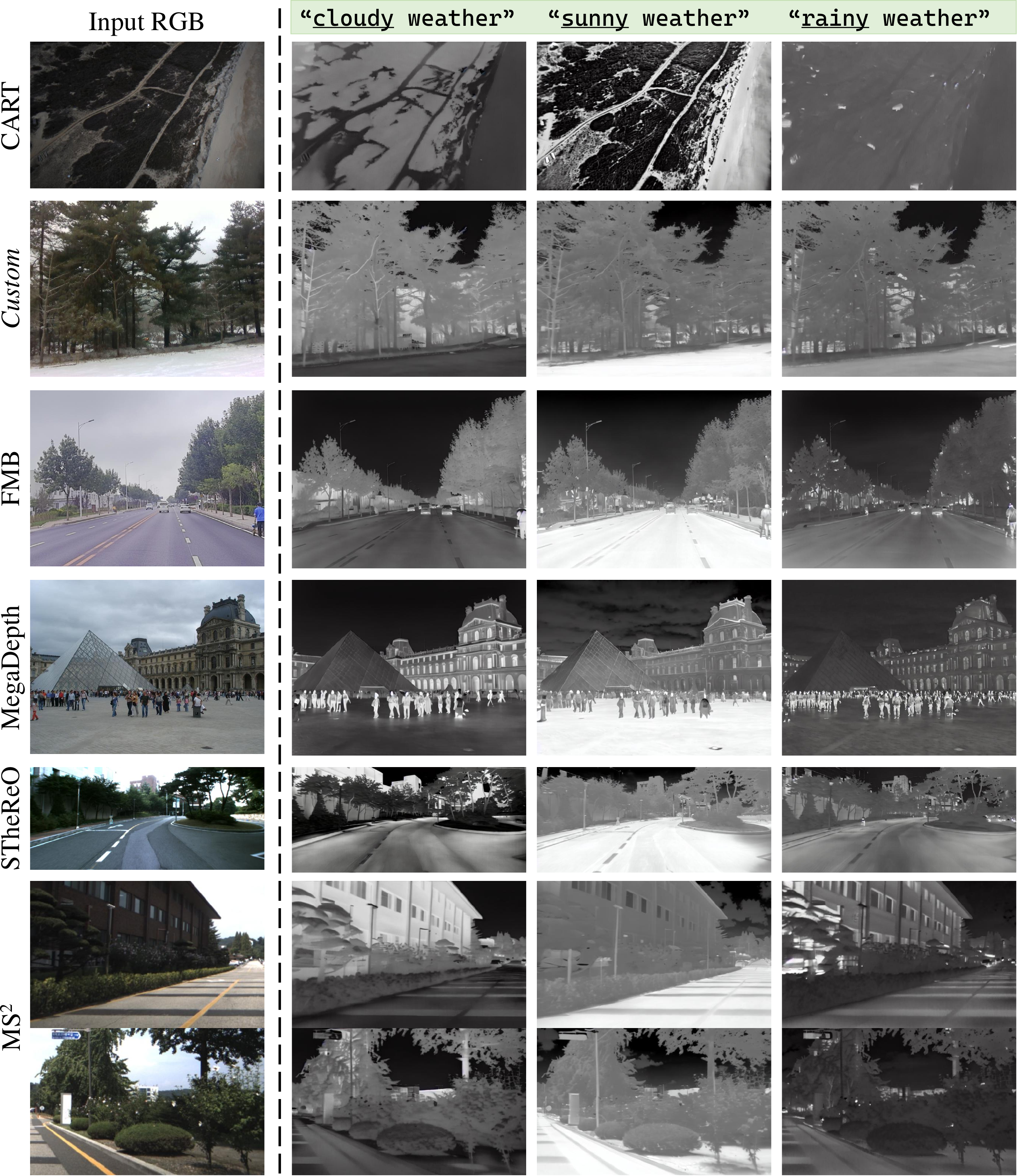}
 \caption{\textbf{Text-guided controllability} Qualitative results of \textit{cloudy, sunny, rainy} weather on different datasets: CART \cite{lee2024caltech-cart}, custom dataset, FMB \cite{fmb}, MegaDepth \cite{li2018megadepth}, MS2 \cite{ms2}, STheReO \cite{sthereo}. MS2 results were horizontally split due to its long aspect ratio.}
  \label{fig:appendix_text_control}
\end{figure*}

\subsection{Controllability Qualitative Results}
\subsubsection{Text-guided Control Qualitative Results}

In \cref{fig:appendix_text_control}, we present additional qualitative examples of text-guided controllability. For each RGB input, we generate translations conditioned on different weather prompts across various public benchmark datasets and a custom snowy scene, all in a zero-shot setting with diverse fields of view, seasons, and locations. We observe that the translation consistently adapts to the requested environmental condition. For example, in the custom snowy image, the apparent temperature of the ground terrain changes with the weather prompt, a behavior that is absent in existing RGB-to-TIR models. This controllability is enabled by the user-interactive TherA-VLM, which allows us to inject prior thermal conditions into the prompt that would not be recoverable from the RGB image alone.

\newpage
\subsubsection{Reference-guided Control Qualitative Results}

\begin{figure*}
 \centering
 \includegraphics[width=0.85\textwidth]{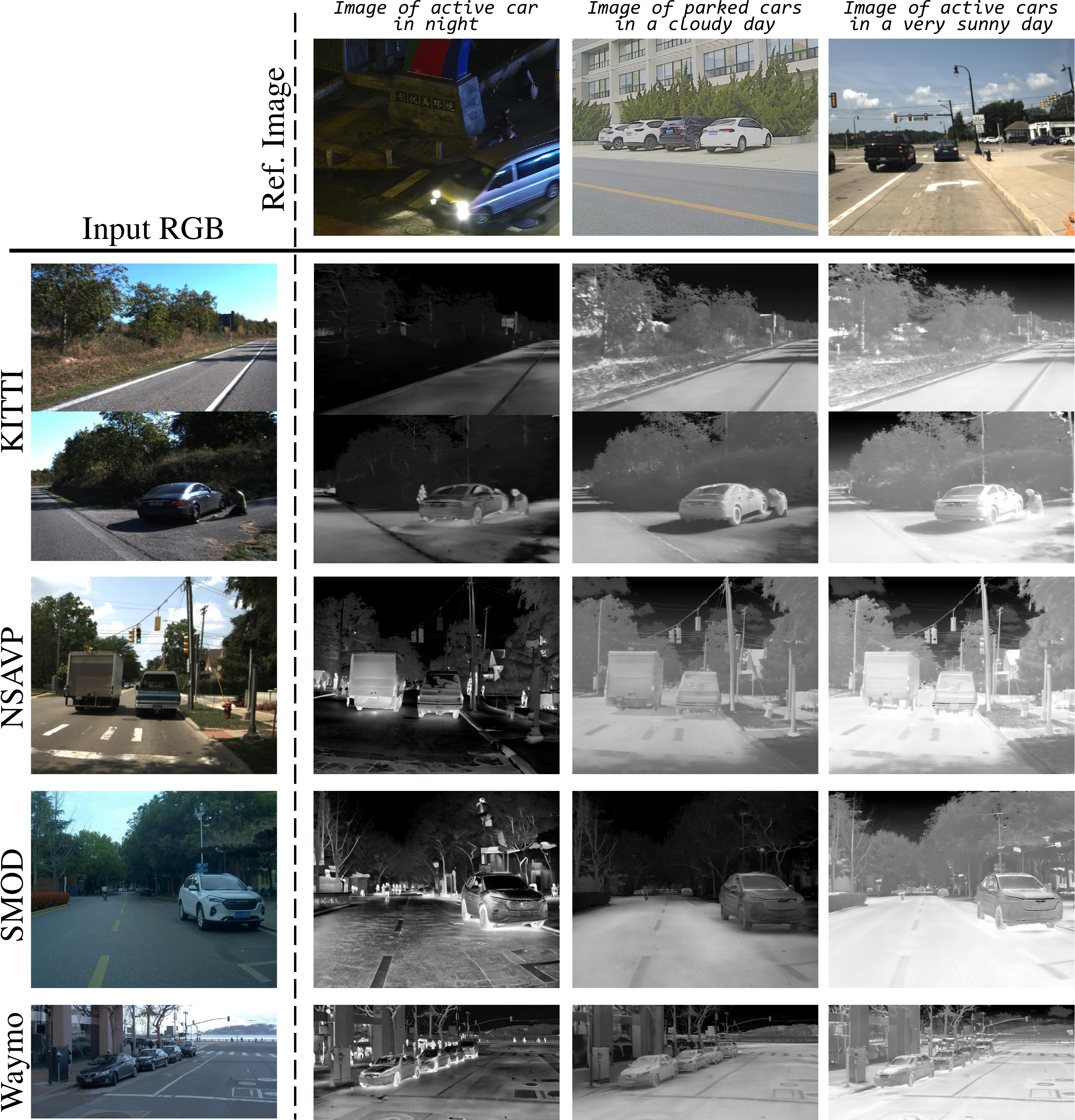}
 \caption{\textbf{Reference-guided controllability} Qualitative results of reference-guided control with reference RGB images from LLVIP \cite{llvip}, M3FD \cite{m3fd}, NSAVP \cite{nsavp} on translating images from KITTI \cite{geiger2013visionkitti}, NSAVP \cite{nsavp}, SMOD \cite{sjtu}, Waymo \cite{sun2020waymo}. KITTI results were horizontally split due to its long aspect ratio.}
  \label{fig:appendix_reference_control}
  \vspace{+3mm}
\end{figure*}

We also provide additional examples of reference-guided controllability in \cref{fig:appendix_reference_control}. For each input RGB image, we condition the translator on a reference RGB depicting an active car at night, parked cars on a cloudy day, or active cars on a sunny day. The translated TIR images adapt global appearance and object-level heat patterns to follow the thermal characteristics implied by the reference. With the night-time active-car reference, vehicles become strong hotspots against a cooler background; with the parked-car reference, vehicles appear cooler and closer to the background; with the sunny-day reference, road surfaces and buildings brighten from solar loading while active vehicles remain highlighted. These results indicate that conditioning through TherA-VLM allows the diffusion model to inherit scene-wise illumination and object-wise emission cues from the reference image, even when the input and reference come from different datasets.

\clearpage

\subsection{Zero-shot Qualitative Results}
\label{sec:zeroshot_qualitative}

\begin{figure*}[t]
 \centering
 \includegraphics[width=\textwidth]{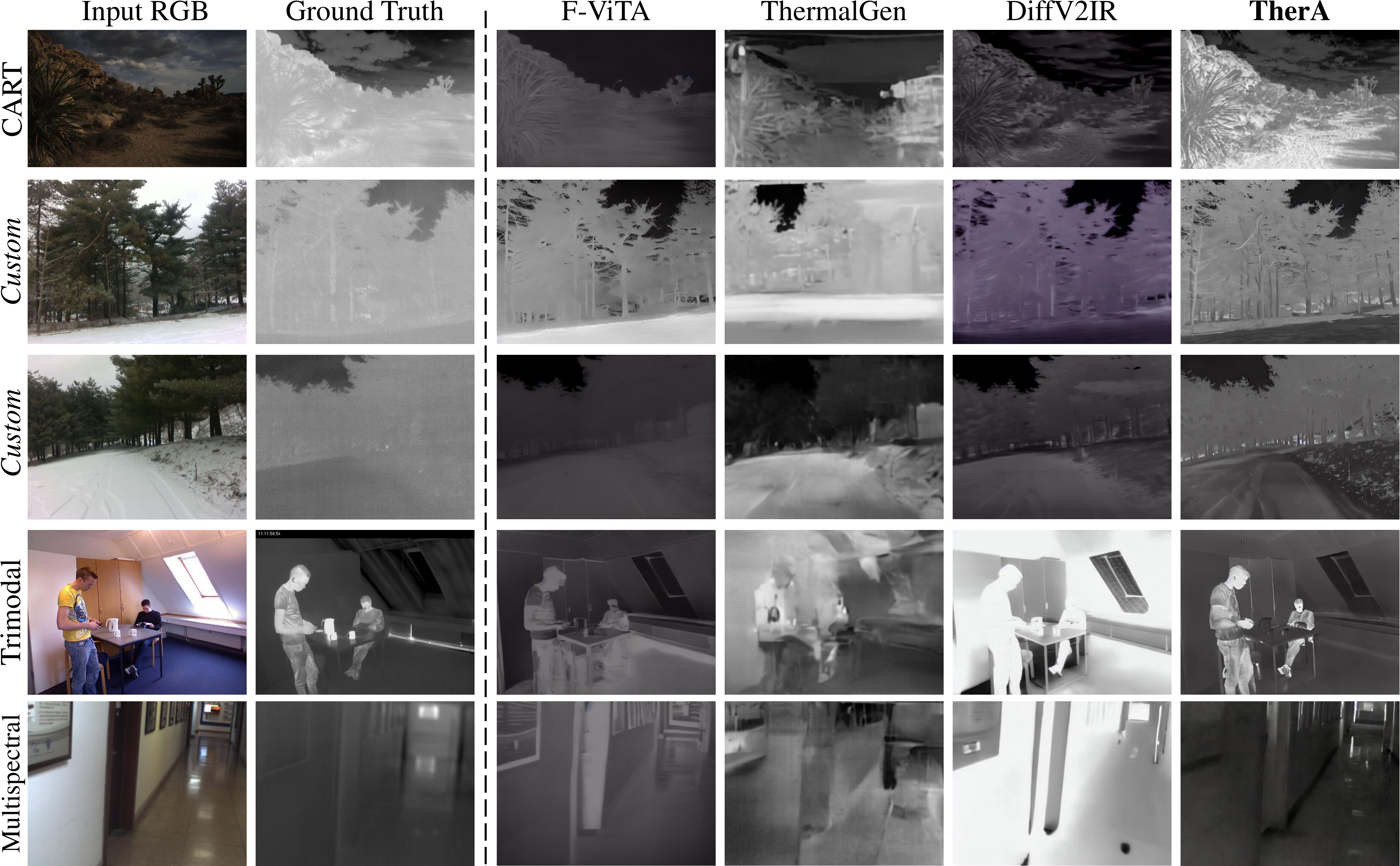}
 \caption{\textbf{Zero-shot Results} Qualitative results on RGB-TIR datasets: CART \cite{lee2024caltech-cart}, custom dataset, Trimodal \cite{trimodal}, Multispectral Motion Dataset \cite{multispectral_motion_dataset}.}
  \label{fig:appendix_zeroshot_yesgt}
\vspace{+5mm}
\end{figure*}

\begin{figure*}[t]
 \centering
 \includegraphics[width=\textwidth]{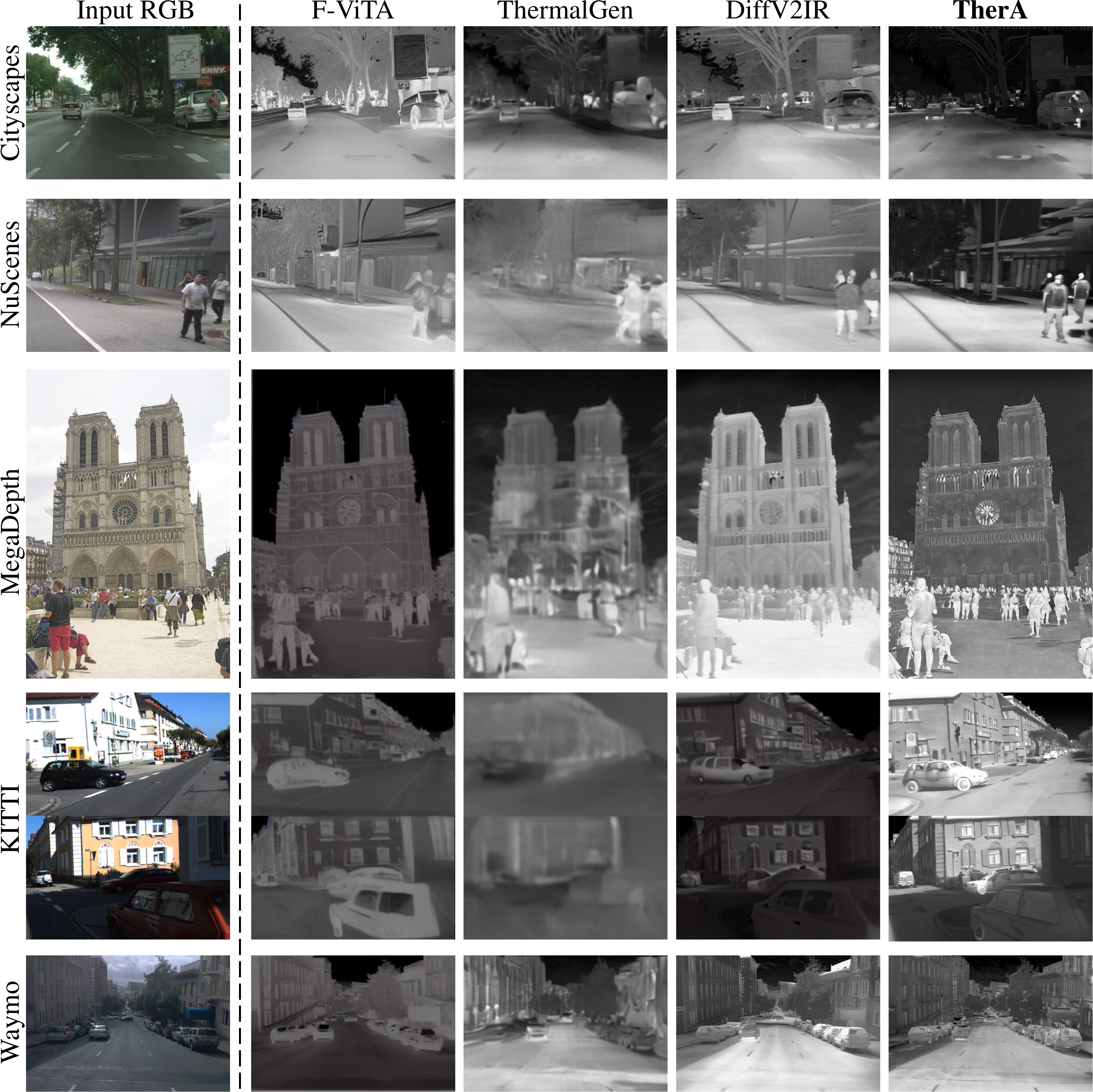}
 \caption{\textbf{Zero-shot Results} Qualitative results on RGB-only datasets: Cityscapes \cite{cordts2016cityscapes}, NuScenes \cite{caesar2020nuscenes}, MegaDepth \cite{li2018megadepth}, KITTI \cite{geiger2013visionkitti}, Waymo \cite{sun2020waymo}. KITTI results were horizontally split due to its long aspect ratio.}
  \label{fig:appendix_zeroshot_nogt}
\end{figure*}

We show additional translation results of TherA in \cref{fig:appendix_zeroshot_yesgt} and \cref{fig:appendix_zeroshot_nogt}.

In \cref{fig:appendix_zeroshot_yesgt}, we compare zero-shot translations on paired RGB--TIR datasets where ground-truth TIR is available (CART \cite{lee2024caltech-cart}, custom scenes, Trimodal \cite{trimodal}, Multispectral Motion \cite{multispectral_motion_dataset}). Baseline models (F-ViTA, ThermalGen, DiffV2IR) frequently exhibit artifacts and thermally implausible outputs. For instance, in the snowy examples (second and third rows), F-ViTA produces unnaturally hot ground despite visible snow cover, and ThermalGen introduces severe ghosting and background bleeding. DiffV2IR yields more stable results but still struggles in indoor scenes, where walls and furniture appear over-heated and fine structural details are washed out. In contrast, TherA produces TIR images that more closely match the ground truth, preserving the relative temperature ordering between snow and vegetation, people and background, and corridor structures.

\newpage

In \cref{fig:appendix_zeroshot_nogt}, we evaluate true zero-shot generalization on RGB-only datasets (Cityscapes, NuScenes, MegaDepth, KITTI, Waymo). Without any paired supervision from these domains, most baselines show a noticeable degradation in quality: contrast is reduced, semantic boundaries become blurry, and expected thermal behavior is often violated (e.g., distant parked vehicles rendered as overly bright moving hotspots, or façades appearing warmer than sunlit roads). While DiffV2IR remains the strongest baseline, its translations are still softer and less physically consistent than those of TherA. TherA maintains sharper details and more realistic thermal distributions, such as cooler skies, thermally structured building façades, and appropriately warm roads and vehicles under traffic scenes.

\newpage
\null
\newpage
\null
\newpage

\section{Ablation Studies}
\label{sec:ablation_studies}

\subsection{CFG Scales}
\label{sec:ablation_cfg}

\begin{table}[!h]
\centering
\caption{Ablation results for \ac{CFG} scales. 
Best results are highlighted in \textbf{bold} and second best are \underline{underlined}.}
\resizebox{1.0\columnwidth}{!}{%
\begin{tabular}{c|c|cccc}
\toprule
\multicolumn{2}{c|}{\textbf{CFG Scale}} & \multirow{2}{*}{\textbf{PSNR(↑)}} & \multirow{2}{*}{\textbf{SSIM(↑)}} & \multirow{2}{*}{\textbf{FID(↓)}} & \multirow{2}{*}{\textbf{LPIPS(↓)}} \\
\cmidrule{1-2}
Thermal VLM & Image control & & & & \\
\midrule
\multicolumn{1}{c|}{\multirow{3}{*}{0.5}} & 0.5 & 19.46 & \underline{0.66} & \underline{87.17} & 0.21 \\
\multicolumn{1}{c|}{} & 1.5 & 18.67 & 0.64 & 97.60 & 0.23 \\
\multicolumn{1}{c|}{} & 2.5 & 18.29 & 0.62 & 102.69 & 0.24 \\ \midrule
\multicolumn{1}{c|}{\multirow{3}{*}{1.5}} & 0.5 & \textbf{19.54} & \textbf{0.67} & \textbf{87.08} & \textbf{0.21} \\
\multicolumn{1}{c|}{} & 1.5 & 18.78 & 0.64 & 96.55 & 0.22 \\
\multicolumn{1}{c|}{} & 2.5 & 18.38 & 0.62 & 101.78 & 0.23 \\ \midrule
\multicolumn{1}{c|}{\multirow{3}{*}{2.5}} & 0.5 & \underline{19.52} & 0.66 & 87.68 & \underline{0.21} \\
\multicolumn{1}{c|}{} & 1.5 & 18.86 & 0.64 & 96.36 & 0.22 \\
\multicolumn{1}{c|}{} & 2.5 & 18.46 & 0.62 & 100.93 & 0.23 \\ \bottomrule
\end{tabular}%
}
\label{translation_results_ablation_cfg}
\end{table}

We analyze the impact of the dual \ac{CFG} scales for image guidance ($s_V$) and VLM guidance ($s_S$) in \Cref{translation_results_ablation_cfg}.
The results reveal a clear and significant trend.
First, increasing the image guidance ($s_V$) consistently degrades performance across all metrics. For instance, at $s_S=1.5$, increasing $s_V$ from 0.5 to 2.5 drops PSNR from 19.54 to 18.38 and worsens FID from 87.08 to 101.78.
Conversely, the VLM guidance ($s_S$) demonstrates an optimal ratio with all metrics peaking at $s_S=1.5$.

This analysis shows that optimal translation is achieved by minimizing reliance on the raw RGB latent ($s_V=0.5$) and assigning more weights on the thermally aware VLM guidance ($s_S=1.5$). This confirms TherA-VLM's thermal embedding provides a more robust and physically grounded signal than the RGB latent alone. We use $s_V=0.5$ and $s_S=1.5$ for all experiments.

\subsection{Downstream Evaluation}
\label{sec:ablation_downstream}
\subsubsection{Thermal Image Segmentation}
\label{sec:ablation_segmentation}

To assess the practical utility of our translations, we evaluate their impact on downstream thermal semantic segmentation. We generate pseudo-\ac{TIR} data by translating Cityscapes \cite{cordts2016cityscapes} with each RGB-to-\ac{TIR} model and train SegFormer \cite{xie2021segformer} on the resulting images. We then evaluate this model in two controlled settings: (1) \emph{Zero-shot} transfer to FMB \cite{fmb} and MFNet \cite{ha2017mfnet}, and (2) \emph{Fine-tuned}, where the same SegFormer is further trained on the real FMB/MFNet training sets. Apart from the translation model used to synthesize the Cityscapes pseudo-\ac{TIR} data, all factors (architecture, training schedule, and labeled target data) are kept fixed. We also include a ``Real TIR'' baseline trained only on the real FMB/MFNet training images, without any synthetic pre-training. For training, we kept all hyperparameters identical to the original implementation, except that we used a batch size of 4 and trained for 100 epochs. 

As summarized in \cref{segmentation_results}, TherA consistently yields the best segmentation performance. In the zero-shot setting, TherA achieves the highest mIoU on both datasets, reaching 27.14 on FMB and 30.43 on MFNet; on MFNet, this corresponds to a +5.05 absolute mIoU gain over the next-best model (ThermalGen). After fine-tuning on real data, models pre-trained on TherA's pseudo-\ac{TIR} data still retain a clear advantage: our approach obtains 52.05 mIoU on FMB and 68.60 mIoU on MFNet, outperforming all other translation baselines and even the Real TIR baselines trained from scratch (42.04 and 57.09 mIoU, respectively).

These results suggest that, under a controlled segmentation pipeline, TherA produces pseudo-\ac{TIR} data that is both more informative and more physically plausible for downstream learning than alternative RGB-to-\ac{TIR} translation models, leading to stronger zero-shot performance and more effective fine-tuning on limited real-world thermal data. In particular, the fact that TherA pre-training even surpasses training directly on real \ac{TIR} indicates that our pseudo-\ac{TIR} images encode thermal contrast patterns that are easier to learn from while remaining consistent with the statistics of real thermal imagery.

\begin{table}[t]
\centering
\caption{Quantitative comparison of class-free TIR semantic segmentation \textbf{mIoU results} [\%] on FMB \cite{fmb} and MFNet \cite{ha2017mfnet}
datasets. \textit{Zero-shot}: SegFormer trained on the Cityscapes dataset \cite{cordts2016cityscapes} translated using each model. \textit{Fine-tuned}: The models further fine-tuned on the FMB or MFNet datasets. Best results are highlighted in \textbf{bold} and second best are \underline{underlined}.}
\resizebox{1.0\columnwidth}{!}{%
\begin{tabular}{l|cc|cc}
\toprule
\multicolumn{1}{c|}{\multirow{3}{*}{Method}} & \multicolumn{2}{c|}{\textbf{FMB} \cite{fmb}} & \multicolumn{2}{c}{\textbf{MFNet} \cite{ha2017mfnet}} \\
\cmidrule(lr){2-3} \cmidrule(lr){4-5}
& Zero-shot & Fine-tuned & Zero-shot & Fine-tuned
\\ 
 \midrule
Real TIR & \textit{N/A} & 42.04 & \textit{N/A} & 57.09 \\
F-ViTA \cite{paranjape2025fvita} & \underline{26.25} & 50.81 & 21.98 & 64.80 \\
ThermalMGen \cite{xiao2025thermalgen} & 23.49 & 50.19 & \underline{25.38} & \underline{66.24} \\
DiffV2IR \cite{ran2025diffv2ir} & 26.17 & \underline{50.88} & 23.75 & 64.38 \\
\textbf{TherA} & \textbf{27.14} & \textbf{52.05} & \textbf{30.43} & \textbf{68.60} \\
\bottomrule
\end{tabular}%
}
\label{segmentation_results}
\end{table}

\subsubsection{RGB--TIR Image Matching}
\label{sec:ablation_matching}

We further study how different translation models affect downstream RGB--\ac{TIR} perception by experimenting with cross-modal image matching. Starting from the RGB-only MegaDepth \cite{li2018megadepth} dataset, we generate pseudo-\ac{TIR} images using each RGB-to-\ac{TIR} translator and train the same LoFTR matcher \cite{sun2021loftr} on the resulting RGB--pseudo-\ac{TIR} pairs. As a reference, we also include a \emph{Baseline (RGB)} model where LoFTR is trained only on original RGB--RGB pairs from MegaDepth (no translation) and then directly applied to RGB--\ac{TIR} matching. In all cases, only synthetic data is used for training; no real thermal images or real RGB--\ac{TIR} pairs from the target dataset are seen during training. We then evaluate the learned matchers \emph{zero-shot} on real RGB--\ac{TIR} pairs from the METU-VisTIR dataset \cite{xoftr-metu-vistir}, reporting pose-estimation AUC at $5^\circ$, $10^\circ$, and $20^\circ$. Similar to our segmentation experiments, all components of the matching pipeline (network architecture, loss, training schedule, and evaluation protocol) are kept fixed; the only varying factor is the translation model used to synthesize the training \ac{TIR} images. For training, we kept all hyperparameters identical to the original implementation, except that we used a batch size of 8 and trained for 25 epochs.

Table~\ref{matching_results} summarizes the results. TherA yields the best performance across all thresholds, achieving AUC@${5^\circ}=14.98$, AUC@${10^\circ}=28.98$, and AUC@${20^\circ}=45.24$, compared to the next-best DiffV2IR with $11.83$, $26.03$, and $43.17$, respectively. In contrast, F\mbox{-}ViTA and ThermalGen provide only marginal gains over the RGB baseline: their AUC scores differ from \emph{Baseline (RGB)} by at most $\sim$2--3 points, and even degrade performance at some thresholds (e.g., F\mbox{-}ViTA at $5^\circ$). This indicates that their pseudo-\ac{TIR} outputs do little to close the RGB--\ac{TIR} modality gap for matching. By comparison, DiffV2IR and especially TherA deliver large improvements over the RGB baseline, showing that physically plausible translation is crucial for learning robust cross-modal correspondences. Because all components of the matching pipeline are held fixed and only the synthetic \ac{TIR} translator is varied, these gains provide further evidence that TherA’s pseudo-\ac{TIR} images encode more physically meaningful cross-modal cues rather than merely improving perceptual image quality.


\begin{table}[h]
\centering
\caption{Quantitative comparison of RGB-TIR image matching AUC results [\%] on METU\_VisTIR \cite{xoftr-metu-vistir} dataset. Best results are highlighted in \textbf{bold} and second best are \underline{underlined}.}
\label{matching_results}
\resizebox{\columnwidth}{!}{%
\begin{tabular}{l|ccc}
\hline
\multicolumn{1}{c|}{\multirow{2}{*}{Method}}          & \multicolumn{3}{c}{$\textbf{LoFTR}$}                                                             \\ \cline{2-4} 
\multicolumn{1}{c|}{}                                 & AUC@$5^\circ$                  & AUC@$10^\circ$                 & AUC@$20^\circ$                 \\ \hline
Baseline (RGB)                                           & 5.44                           & 12.58                          & 24.28                          \\
F-ViTA \cite{paranjape2025fvita}     & 4.93                           & 12.73                          & 25.32                          \\
ThermalGen \cite{xiao2025thermalgen} & 5.81                           & 14.24                          & 27.65                          \\
DiffV2IR \cite{ran2025diffv2ir}      & \underline{11.83} & \underline{26.03} & \underline{43.17} \\
\textbf{TherA}                                        & \textbf{14.98}                 & \textbf{28.98}                 & \textbf{45.24}                 \\ \hline
\end{tabular}%
}
\end{table}

\subsection{Comparison with Reference-guided Models}
\label{sec:ablation_referenceguided}

\begin{figure*}[t]
 \centering
 \includegraphics[width=\textwidth]{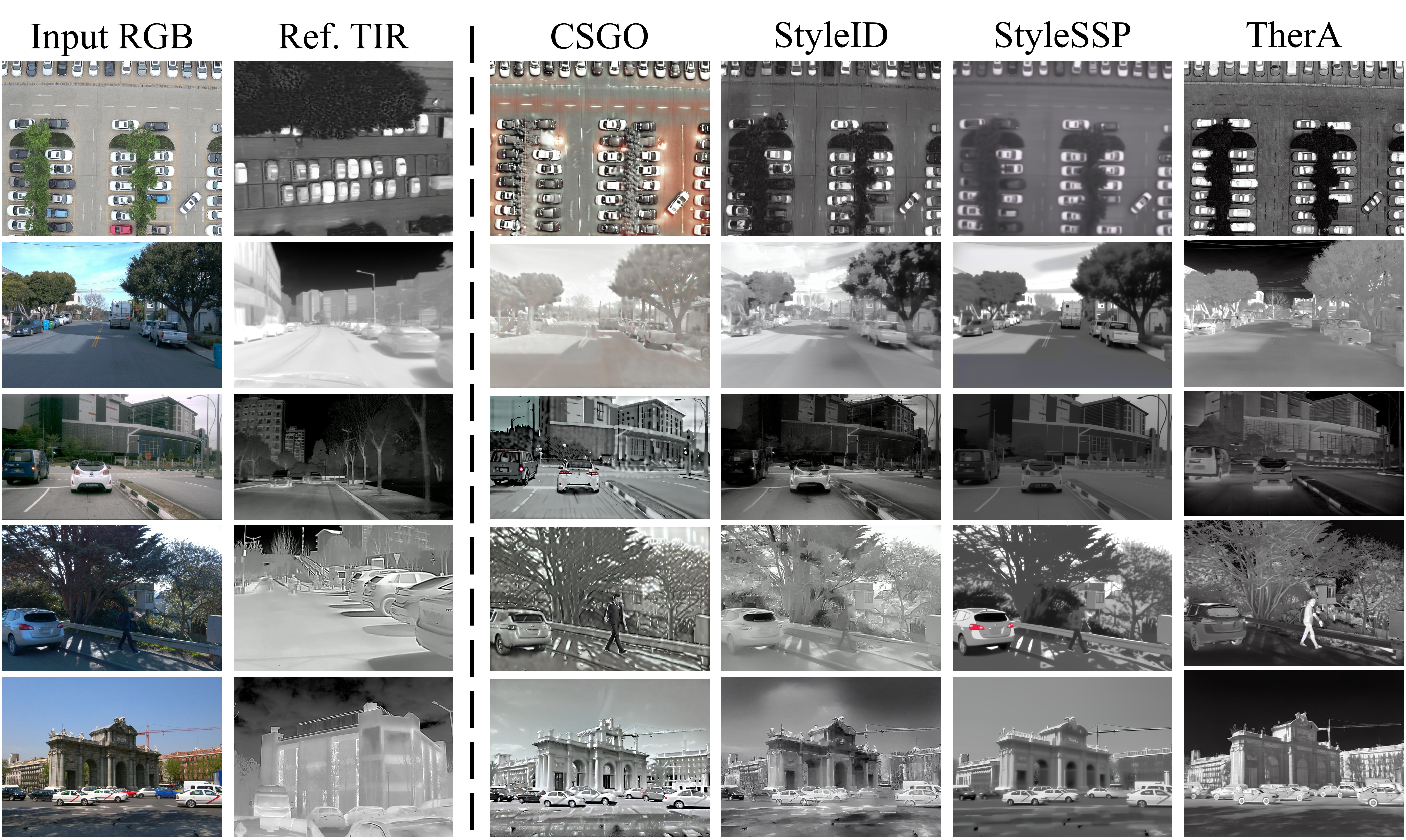} 
 \caption{
  \textbf{Reference-guided translation comparison.} Qualitative comparison of existing reference-guided translation models and TherA. Source images taken from popular RGB benchmarks (from the top: Visdrone \cite{zhu2021detection_visdrone}, Waymo \cite{sun2020waymo}, NuScenes \cite{caesar2020nuscenes}, Waymo \cite{sun2020waymo}, and MegaDepth \cite{li2018megadepth})}. 
 \label{fig:ref-guided-translation}
\end{figure*}

We evaluated the reference-guided image translation performance of TherA against state-of-the-art example-guided image-to-image translation models, including CSGO~\cite{xing2024csgo}, StyleID~\cite{chung2024styleid}, and StyleSSP~\cite{xu2025stylessp}. For all three baselines, we use the official implementations and publicly released pre-trained weights. For CSGO, we adopt the reference-guided setting by feeding the RGB input as the content image and the reference TIR frame as the style image. For StyleID and StyleSSP, we evaluate them in their recommended training-free configuration, using the RGB frame as the source and the TIR frame as the style exemplar. All methods receive exactly the same RGB–TIR pairs and reference images; only slight exception exists for TherA which leverages the RGB reference image as the reference condition in contrast to other methods which leverage TIR images as the reference image for translation. But in the end, only the translation module is varied while the evaluation protocol is kept fixed.

As shown in \cref{fig:ref-guided-translation}, baseline methods struggle to effectively bridge the modality gap between the visible and thermal domains. Specifically, CSGO exhibits domain-inconsistent artifacts and residual chromatic noise, failing to adequately suppress RGB color information. StyleID and StyleSSP suffer from over-smoothing and fail to capture distinct thermal physics. As a result, these baselines often fail to consistently map semantic classes to appropriate thermal intensities, frequently rendering heat-emitting objects such as pedestrians or vehicle engines with low contrast.

In contrast, TherA produces the most perceptually realistic thermal imagery: it successfully disentangles RGB texture from structure, accurately synthesizes thermal infrared appearance for distinct semantic objects, and faithfully adheres to the textural style and polarity of the reference guidance. For example, in the first row, all other methods incorrectly depict the white car as one of the hottest objects; in practice, because white surfaces tend to reflect incident thermal infrared more than darker, more emissive materials, they should appear relatively cooler, and TherA is the only method that captures this behavior. Moreover, TherA is the only model that consistently identifies pedestrians with physically plausible heat-emitting characteristics. Since all evaluations are performed in a zero-shot setting on popular benchmark datasets, these results not only demonstrate TherA’s strong generalization ability but also establish TherA as the first example-guided translation module explicitly specialized for RGB-to-TIR translation.

We do not fine-tune any of these baseline models on thermal data, so they fully benefit from large-scale RGB pre-training, yet the baseline reference-guided models still fail to recover physically meaningful thermal appearance. This demonstrates that RGB-to-TIR translation cannot be solved by generic appearance-based style transfer alone, and instead requires a thermal-aware model that explicitly accounts for semantic structure and imaging physics.

%

\section{Limitations and Failure Cases}
\label{sec:limitations}

\paragraph{Dependence on RGB visibility and VLM conditioning.}
TherA generates pseudo-\ac{TIR} images by conditioning a diffusion model on TherA-VLM. As a result, failure of either component can lead to suboptimal translations.
We observe that errors are most common in frames with severely degraded RGB images, such as strong motion blur, almost completely dark scenes, or saturated/overexposed regions where scene geometry and texture are barely visible.
In such cases TherA-VLM may produce incomplete or noisy descriptions and the diffusion model has little visual guidance, which can result in unrealistic or unstable \ac{TIR} predictions.
Our prompting explicitly instructs the VLM to abstain from low-confidence statements, and user-side control can partially mitigate this by editing the instruction using text-guided or reference image-guided control, but the method fundamentally assumes that the RGB image contains sufficient visual information.

\paragraph{Relative temperature representation.}
Our current system operates on normalized thermal imagery: pixel intensities are normalized and only encode local temperature ordering and contrast, not absolute radiometric values.
Consequently, TherA is not suitable for applications that require metric temperature estimation, emissivity calibration, or strict compliance with a full radiometric chain.
Instead, our translations are designed to preserve qualitative thermal structure (relative hot/cold patterns, material- and scene-dependent contrast) and to support perception tasks such as recognition, matching, and segmentation.
Extending the framework to calibrated radiometric data and explicitly modeling absolute temperature remains an important direction for future work.

\paragraph{Scope of active/passive controllability.}

\begin{figure}
 \centering
 \includegraphics[width=\columnwidth]{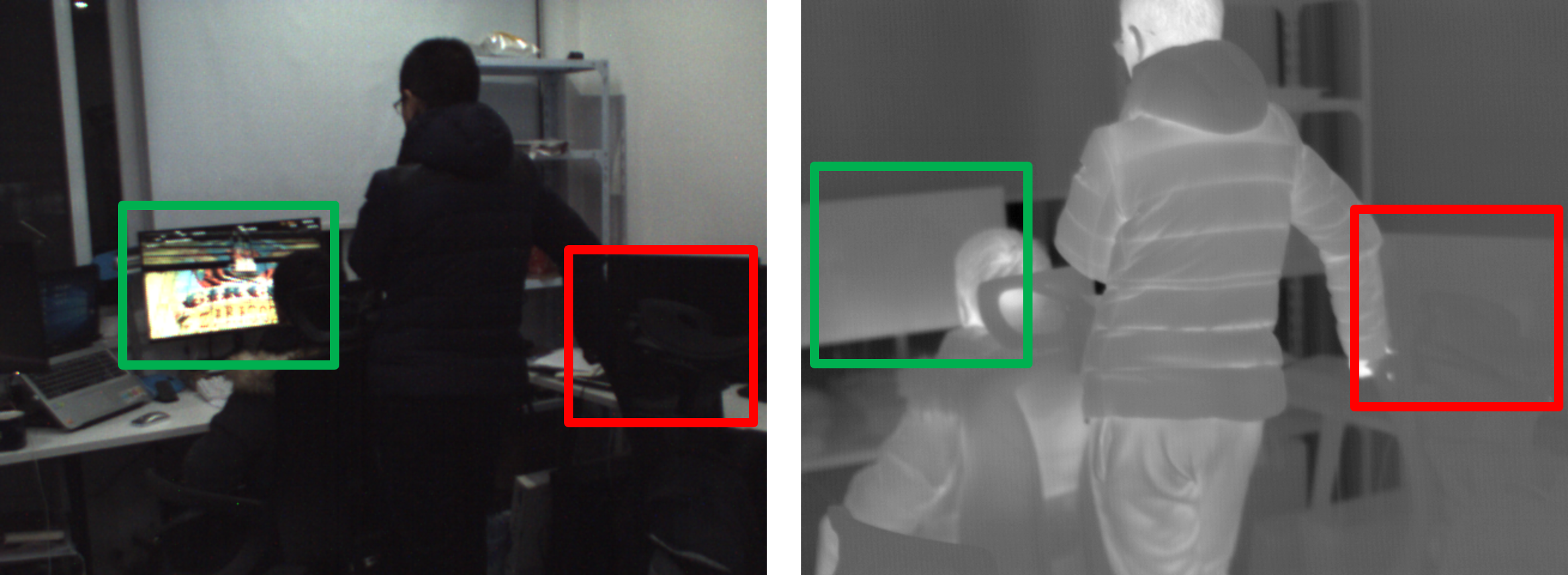}
 \caption{Limitation of controllable active/passive editing for indoor electronics. In the RGB image (left), a powered-on monitor (green box) and a powered-off monitor (red box) are clearly distinguishable. In the corresponding real TIR image (right), however, the temperature contrast between the two displays is subtle, making it difficult for TherA to produce a visually striking on/off effect for such objects.}
  \label{fig:monitor-failure}
   \vspace{-5mm}
\end{figure}

While TherA can toggle the active/passive state of objects via reference image guidance, the visual effect is most pronounced for objects that exhibit strong and spatially extended heat emission, such as vehicles and human bodies in outdoor scenes.
For other object categories, especially small indoor electronics (e.g., monitors, laptops) or surfaces with weak thermal contrast, the change in the real \ac{TIR} domain is often subtle.
Figure~\ref{fig:monitor-failure} illustrates such a case: the RGB image contains both an active display and an inactive monitor, yet the corresponding thermal image shows only a modest intensity difference between them.
Our translations reflect this limited contrast rather than a dramatic on/off switch, and indoor scenes with nearly uniform ambient temperature similarly offer limited room for controllable variation.
This bias stems from the underlying training data distribution and highlights that our controllable attributes are currently most reliable for large-scale outdoor structures and vehicles, rather than all possible heat sources.

Looking ahead, we expect that stronger thermal-aware VLM backbones and joint end-to-end fine-tuning of the VLM and diffusion components on more diverse large-scale RGB--\ac{TIR} data will further reduce these failure modes and improve robustness in challenging illumination regimes.


\end{document}